**Explainable AI Methods for Neuroimaging: Systematic Failures of Common Tools, the Need for Domain-Specific Validation, and a Proposal for Safe Application**

Nys Tjade Siegel[1, 2], James H. Cole[3, 4, 5], Mohamad Habes[6], Stefan Haufe[7, 8, 9, 10], Kerstin Ritter[1, 2, 11,*], Marc-André Schulz[1, 2,*,+]

[1] Department of Psychiatry and Neurosciences, Charité – Universitätsmedizin Berlin (corporate member of Freie Universität Berlin, Humboldt-Universität zu Berlin, and Berlin Institute of Health), Berlin, Germany
[2] Department of Machine Learning, Hertie Institute for AI in Brain Health, University of Tübingen, Germany
[3] UCL Hawkes Institute, Faculty of Engineering, University College London, UK
[4] Department of Computer Science, University College London, UK
[5] Dementia Research Centre, Queen Square Institute of Neurology, University College London, UK
[6] Neuroimage Analytics Laboratory and the Biggs Institute Neuroimaging Core, Glenn Biggs Institute for Alzheimer's and Neurodegenerative Diseases, University of Texas Health Science Center at San Antonio, USA
[7] Bernstein Center for Computational Neuroscience, Berlin, Germany
[8] Technische Universität Berlin, Berlin, Germany
[9] Physikalisch-Technische Bundesanstalt, Berlin, Germany
[10] Department of Neurology, Charité – Universitätsmedizin Berlin (corporate member of Freie Universität Berlin, Humboldt-Universität zu Berlin, and Berlin Institute of Health), Berlin, Germany
[11] Tübingen AI Center, Tübingen, Germany
* shared senior authors
*correspondence to marc-andre.schulz@charite.de

## Abstract

Trustworthy interpretation of deep learning models is critical for neuroimaging applications, yet commonly used Explainable AI (XAI) methods lack rigorous validation, risking misinterpretation. We performed the first large-scale, systematic comparison of XAI methods on ~45,000 structural brain MRIs using a novel XAI validation framework. This framework establishes verifiable ground truth by constructing prediction tasks with known signal sources - from localized anatomical features to subject-specific clinical lesions - without artificially altering input images. Our analysis reveals systematic failures in two of the most widely used methods: GradCAM consistently failed to localize predictive features, while Layer-wise Relevance Propagation generated extensive, artifactual explanations that suggest incompatibility with neuroimaging data characteristics. Our results indicate that these failures stem from a domain mismatch, where methods with design principles tailored to natural images require substantial adaptation for neuroimaging data. In contrast, the simpler, gradient-based method SmoothGrad, which makes fewer assumptions about data structure, proved consistently accurate, suggesting its conceptual simplicity makes it more robust to this domain shift. These findings highlight the need for domain-specific adaptation and validation of XAI methods, suggest that interpretations from prior neuroimaging studies using standard XAI methodology warrant re-evaluation, and provide urgent guidance for practical application of XAI in neuroimaging.

## Introduction

Deep learning models are increasingly applied in neuroimaging analyses, where they promise advances in disease classification, biomarker discovery, and the study of brain structure and function (Isensee et al., 2021; Litjens et al., 2017). However, the clinical translation and scientific utility of these models are fundamentally limited by their "black box" nature (Kelly et al., 2019; Rudin, 2019). For high-stakes decisions in healthcare and robust neuroscientific inference, simply knowing a model's prediction is insufficient; understanding *why* the model arrived at that prediction - its underlying reasoning - is critical for building trust, ensuring safety, enabling regulatory approval, and generating genuine insight (Holzinger et al., 2019; Muehlematter et al., 2021).

Explainable Artificial Intelligence (XAI) methods are designed to address this interpretability gap, typically by generating attribution or saliency maps that highlight input features - in this context, brain regions - purportedly driving a model's decision (Gilpin et al., 2018; Montavon et al., 2018). Methods like Gradient-weighted Class Activation Mapping (GradCAM) (Selvaraju et al., 2017), Layer-wise Relevance Propagation (LRP) (Bach et al., 2015), and Guided Backpropagation (Springenberg et al., 2014) are increasingly applied in neuroimaging research (Böhle et al., 2019; Eitel et al., 2019; Siegel et al., 2025). Yet, this adoption often outpaces rigorous validation. Concerningly, applying different established XAI methods to the *same* well-performing deep learning model analyzing the *same* neuroimaging data can yield contradictory or mutually exclusive explanations (Fig. 2, Supplementary Material SM-D1), raising profound questions about their reliability in this specific domain (cf. Adebayo et al., 2018; Kindermans et al., 2019).

This lack of reliability stems from a validation gap. Evaluating explanation methods ideally requires ground truth - knowing what features the model *truly* relied upon - which is inherently unavailable for complex models learning intricate patterns (Bommer et al., 2024; Doshi-Velez & Kim, 2017; Yang & Kim, 2019). Without such ground truth, it becomes impossible to distinguish between genuinely incorrect explanations and those that truthfully reflect a model's reliance on shortcuts (Lapuschkin et al., 2019) or spurious correlations (Wang et al., 2023). Despite this major limitation, many studies have, in practice, relied on evaluating whether an explanation "looks plausible" (Tjoa & Guan, 2021). To address this issue, some researchers have attempted to approximate ground truth in natural images - for example, by using object segmentation masks (Kohlbrenner et al., 2020; Pahde et al., 2022; Y. Zhang et al., 2023). These approaches are inadequate for the neuroimaging domain, however, due to fundamental differences in the data: strong spatial correlations, lack of canonical objects, the prevalence of subtle and distributed features rather than sharp edges, and significant inter-subject variability (Marek et al., 2022; Mechelli et al., 2005; Schulz et al., 2020). Existing evaluations of XAI in neuroimaging are scarce and limited by scale or unrealistic modification to the input images (e.g. Budding et al., 2021; Hofmann et al., 2022; Oliveira et al., 2024).

Here, we address this gap by performing the first large-scale, systematic comparison and validation of common XAI methods for structural neuroimaging. We introduce and apply a novel XAI validation framework using data from approximately 45,000 UK Biobank T1-weighted and T2 FLAIR MRI scans. This framework enables objective assessment against verifiable ground truth across a spectrum of increasing complexity - from precisely localized anatomical features to clinically relevant, subject-specific distributed patterns - crucially, *without* artificially modifying the input images, thus preserving the natural properties of the data. Applying this framework, we uncover systematic, widespread failures in the most commonly used XAI methods in neuroimaging (GradCAM and LRP; survey on method usage in SM-G), revealing localization failures and artifact generation. We provide strong evidence that these failures arise from a domain mismatch, whereby methods implicitly optimized for natural image statistics do not generalize reliably to neuroimaging data. Importantly, our framework also identifies simpler gradient-based methods, particularly SmoothGrad (Smilkov et al., 2017), as a consistently accurate alternative across the tested scenarios. Our findings may challenge the interpretations drawn from potentially numerous prior studies (cf. SM-G), provide empirical guidance for researchers and clinicians, and establish a robust methodology for validating the trustworthiness of XAI in neuroimaging.

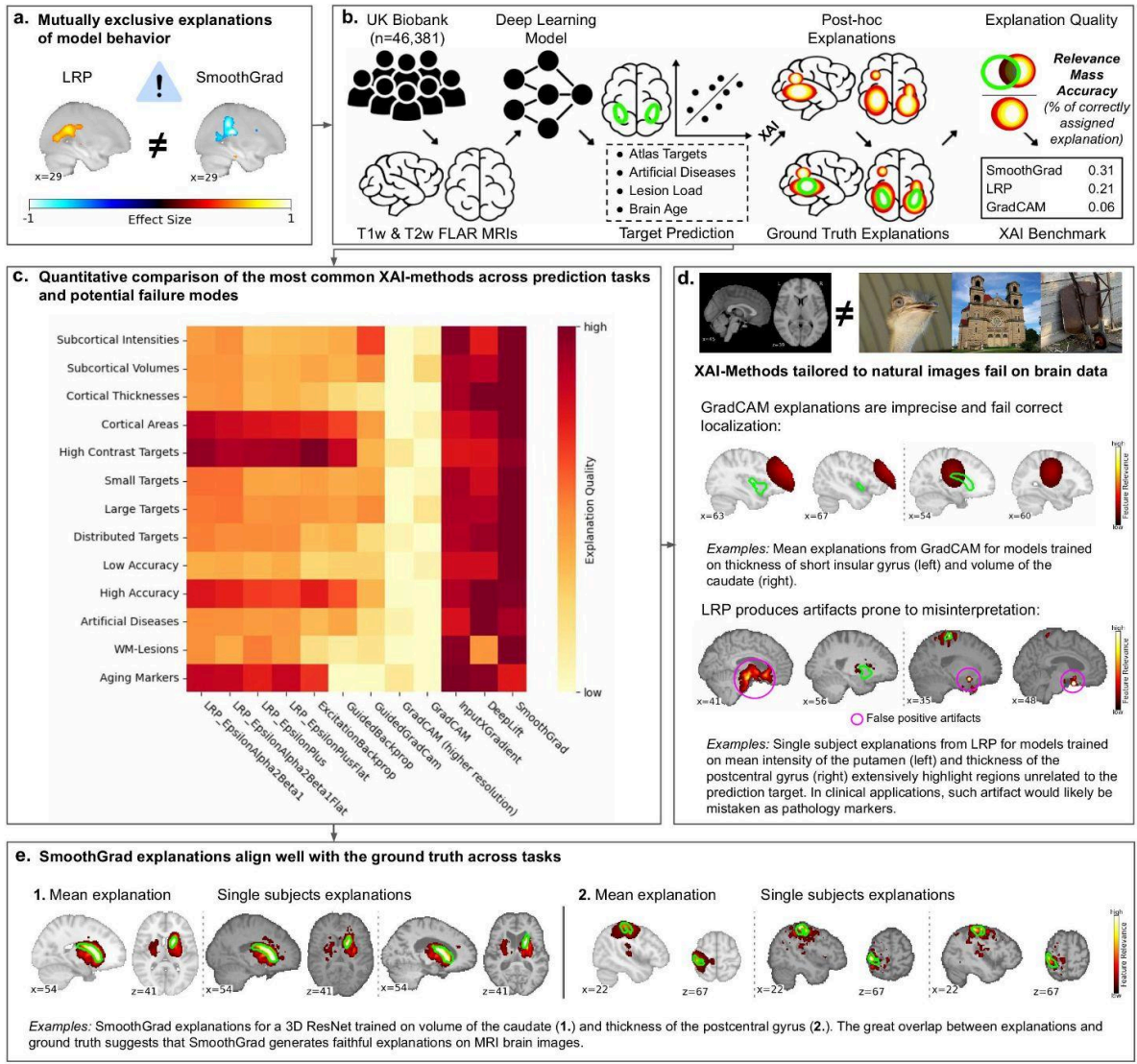

Figure 1: Different XAI methods yield conflicting explanations, necessitating ground truth validation enabled by our framework, which reveals systematic failures and successes of XAI methods. a) Mutually exclusive explanations arise when applying different XAI methods (e.g., LRP vs. SmoothGrad) to the same brain age prediction model analysing patients with MS vs. controls, highlighting the need for objective validation (details in Fig. 2, Supplementary Material SM-D1). b) Overview of the ground-truth validation framework. Structural MRI data from the UK Biobank are used to train a deep learning model on prediction tasks with pre-defined signal sources, including atlas-based targets, artificial diseases, lesion load, and brain age. Post-hoc explanation methods are applied to generate saliency maps, which are then compared to the known ground truth. Explanation quality is quantified using metrics like Relevance Mass Accuracy (RMA; percentage of the explanation signal correctly located within the ground-truth region). c) Systematic evaluation across the framework reveals consistently high explanation quality (RMA) for SmoothGrad but failures for LRP and GradCAM (quantitative results in Supplementary Table ST-2; row-wise min-max scaled scores underlying Fig. 1c in ST-4). d) Common methods LRP and GradCAM exhibit critical failure modes: LRP generates false positive artifacts (examples for Putamen Intensity, Insular Thickness), while GradCAM fails localization (examples for Insular Thickness, Caudate Volume), attributed to a domain mismatch where methods tailored for natural images falter on brain data (details in Fig. 3, Fig. 4). e) In contrast, SmoothGrad consistently and accurately localizes ground truth features across the framework (examples for Caudate Volume, Postcentral Gyrus Thickness; details in Fig. 5).

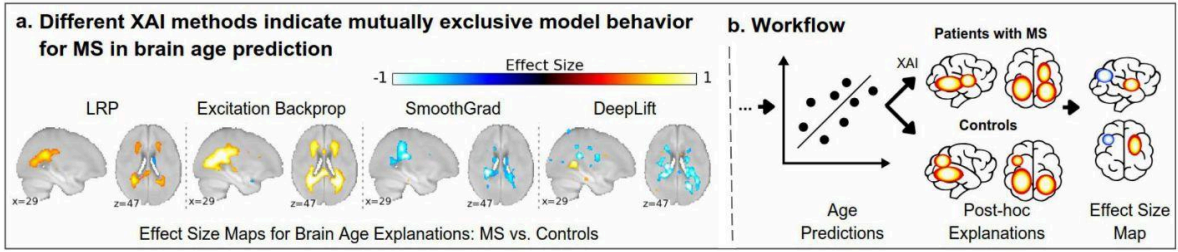

*Figure 2: Different XAI methods applied to the same model and data yield mutually exclusive results. a) Different XAI techniques (e.g., LRP, Excitation Backprop vs. SmoothGrad, DeepLift) applied to the same brain age prediction model yield conflicting insights on model behavior in Multiple Sclerosis (MS). For instance, LRP and Excitation Backprop highlight the ventricles as particularly relevant aging markers in MS, whereas SmoothGrad and DeepLift indicate reduced reliance on ventricular features in this group. (Warm colors = greater explanation mass in patients with MS compared to controls; cold colors = lower explanation mass.) b) Workflow overview: A 3D ResNet-50 model predicted brain age from T2 FLAIR MRI scans (see Fig 1b). Post-hoc explanations (e.g., via LRP, Excitation Backprop, SmoothGrad, DeepLift) were generated for MS patients and matched controls. Effect size maps, masked by FWE-corrected significance ($\alpha = 0.05$), reveal structural differences in the model's explanations across groups (details in SM-D1).*

**Results**

**Evidence of Urgency: Conflicting Explanations Demonstrate the Need for Objective Validation**

The ambiguity inherent in applying unvalidated XAI methods to neuroimaging is starkly illustrated when different techniques analyze the same prediction. We show that, when examining a well-performing deep learning model that predicts brain age - a common neuroimaging biomarker - in patients with multiple sclerosis (MS) versus healthy controls (Brier et al., 2023; Cole et al., 2020; Kaufmann et al., 2019), different XAI methods yield contradictory explanations for the observed brain age differences (Fig. 1a, Fig. 2, SM-D1). One method suggests that the model focuses on ventricles as particularly informative markers of aging in patients with MS, while another suggests that the model disregards the ventricles as features particularly in patients with MS. These opposing explanations suggest mutually exclusive internal decision-making processes in the deep learning model, they cannot both be true. Such conflicting results, generated from the identical model and data, underscore the impossibility of determining the correct interpretation through visual inspection alone and establish the need for objective validation against known ground truth.

**A Multi-Stage Framework for Ground-Truth Validation of XAI in Neuroimaging**

To evaluate XAI methods when the model's "true" reasoning is unknown, we developed a validation framework that establishes verifiable ground truth for explanations. The core principle is to construct prediction tasks where the source of the predictive signal in the input data is known *a priori*, rather than altering the input images themselves, thereby maintaining the natural statistical properties of the data. This framework allows us to systematically assess XAI method reliability against this ground truth across tasks of increasing complexity, using large-scale, unmodified 3D T1- and T2 FLAIR brain MRI data from the UK Biobank (N ≈ 45,000).

The framework comprises four stages of increasing complexity and realism:

**Stage 1: Localized Anatomical Features (Corrected IDPs)**: Our foundational test creates a scenario where the model can derive information about the target only from a single predefined brain region, so that any explanation mass outside that target region can be identified as verifiably spurious. We achieved this by training models to predict corrected Imaging-Derived Phenotypes (cIDPs) - quantitative anatomical measures, like regional volumes, that we processed to be highly specific solely to their corresponding anatomical structure. For a model predicting the cIDP for the caudate nucleus, the anatomical mask of the caudate thus serves as the ground-truth for the explanation. We empirically validated this ground truth; when the target region was computationally removed from the input images, the model's predictive accuracy ($R^2$) dropped to near zero, confirming that the signal was indeed localized as intended (details in SM-A3, Companion Manuscript Table A2).

**Stage 2: Controlled Distributed Patterns ("Artificial Diseases"):** To evaluate whether XAI methods can identify distributed predictive patterns - a key diagnostic challenge - we created "artificial diseases". These are synthetic binary classification targets built by combining cIDPs from multiple, distinct anatomical regions. This design simulates a core clinical problem: detecting concurrent but spatially separate abnormalities, analogous to how conditions like Alzheimer's disease manifest as patterns of atrophy across different brain lobes. This approach establishes an unambiguous ground truth for distributed effects, allowing us to test an XAI method's ability to capture multi-region relevance. (Details in SM-A4).

**Stage 3: Clinically Relevant Distributed Patterns (Lesions):** We train models to predict overall white matter hyperintensity (WMH) lesion load, a clinically significant marker often associated with conditions like stroke, vascular cognitive impairment, and dementia (Debette & Markus, 2010; Habes et al., 2016). For evaluating explanations, the ground truth is derived from subject-specific lesion segmentation masks, representing real-world, clinically meaningful, distributed patterns that vary in location and extent across individuals. This stage provides a crucial test of performance on heterogeneous, pathologically relevant features.

**Stage 4: Complex Biomarker Plausibility (Brain Age):** We utilize brain age prediction - predicting chronological age from brain structure, a task where deep learning excels (Cole & Franke, 2017; Hahn et al., 2022; Siegel et al., 2025). Here, direct spatial ground truth is unavailable. Instead, we perform a literature-driven plausibility check (Thomas et al., 2023; Wang et al., 2023). We generate explanations for the brain age model and compare the spatial distribution of relevance (ranked by brain region) against established anatomical patterns of aging derived from meta-analyses in the neuroimaging literature (Walhovd et al., 2011). This assesses whether explanations align with known, complex biological patterns. (Details in SM-A6).

Within this framework, we trained 3D ResNet-50 models (architecture details in SM-B1; alternative architecture in SM-F4) for each prediction task. The models were able to successfully predict all our targets ($R^2$: 0.27 to 0.88; accuracy: 0.80 to 0.83; full results in Supplementary Table ST-1). We then applied a comprehensive suite of XAI methods, including gradient-based (SmoothGrad, InputxGradient), relevance-based (LRP, using common rule sets), CAM-based (GradCAM), and reference-based (DeepLift) approaches (implementation details in SM-B3). Explanation quality was quantified using established metrics: Relevance Mass Accuracy (RMA; proportion of explanation signal within the ground truth mask), True Positive Rate (TPR; percentage of cases where the target ROI was successfully identified), and False Positive Rate (FPR; how often explanations assigned high relevance to brain regions outside the ground truth mask) (Arras et al., 2022; metric definitions in SA-B4).

### Discovery: Systematic Failures of Common XAI Methods

Applying XAI methods across our validation framework revealed systematic failures in the techniques most commonly employed in the neuroimaging literature: LRP and GradCAM (Fig. 1c, Fig. 3; literature survey on XAI method usage in SM-G).

**GradCAM:** This widely used method (Nazir et al., 2023; van der Velden et al., 2022; SA-G) consistently failed to reliably localize the relevant anatomical features. Quantitatively, GradCAM explanations exhibited low RMA (Fig. 1c, ST-2) and often failed to identify the correct region as most important (low TPR) across numerous tasks (Fig. 3b, TPRs in ST-6). Qualitatively, GradCAM heatmaps were frequently diffuse and misaligned with the ground truth region (Fig. 3b). These localization failures were apparent even for simple, localized IDP targets (e.g., Insular thickness) and persisted in the more complex lesion prediction task, rendering GradCAM unreliable for pinpointing determinative features in neuroimaging data. For a discussion of resolution and layer-level explanation, see Supplementary Material SM-D3.

**LRP:** While sometimes appearing visually sharper than GradCAM, LRP with standard rule sets showed extensive false-positive artifacts. Quantitatively, LRP consistently showed a high FPR (Fig. 3a, ST-7), indicating that regions verifiably unrelated to the prediction task were highlighted as prominent explanations. Qualitatively, this manifested as widespread, often bilateral patterns of activation that extended far beyond the target structure, even for tasks with highly localized ground truth like predicting the intensity of the putamen or the thickness of the short insular gyrus (Fig. 3a). These artifacts, which could easily be misinterpreted as genuine distributed effects in a clinical or research setting, were observed across multiple LRP rule implementations (see SM-D2). Further analysis suggested that LRP might be particularly attuned to image contrast, performing disproportionately well (albeit still producing artifacts) on large shapes, such as the lateral ventricles, compared to the low-contrast subcortical targets (Figure 1c, ST-2).

These failures of the two most prevalent XAI methods in neuroimaging underscore the risk of generating misleading interpretations and potentially invalid conclusions in studies relying on these tools without domain-specific validation.

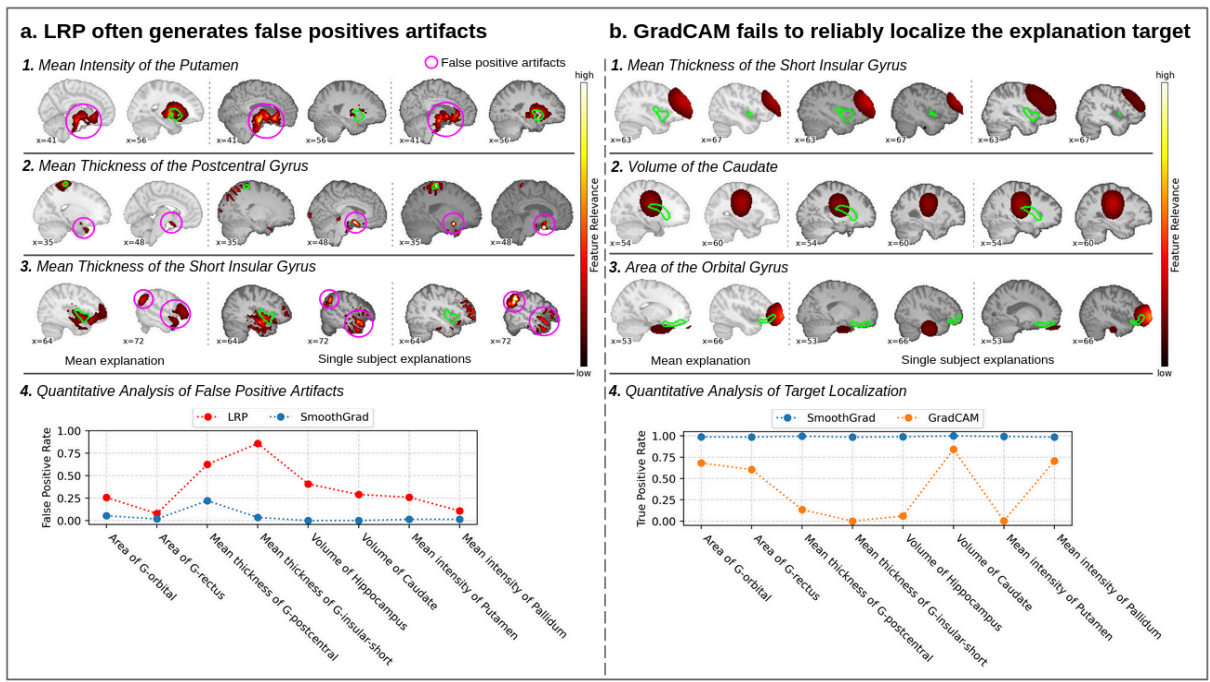

*Figure 3: Common XAI methods LRP and GradCAM exhibit critical failure modes on neuroimaging data. Quantitative analysis (Fig. 1c, SM-C) reveals issues validated qualitatively here. (a) LRP often generates extensive false positive artifacts (high False Positive Rate - FPR). Examples show mean and single-subject explanations for models predicting Putamen intensity, Postcentral Gyrus thickness, and Short Insular Gyrus thickness, where highlighted relevance (yellow/red) extends far beyond the target regions (green outlines), risking misinterpretation (further examples in Supplementary Figure*

*SF-1). (b) GradCAM frequently fails localization (low True Positive Rate - TPR). Examples show mean and single-subject explanations for models predicting Short Insular Gyrus thickness, Caudate volume, and Orbital Gyrus area, where heatmaps are diffuse or misaligned with target regions (green outlines) (further examples in SF-5). (c, d) Quantitative plots summarize these failures across multiple framework tasks, showing LRP's high FPR and GradCAM's low TPR compared to SmoothGrad (full quantitative results in ST-6 (TPR) and ST-7 (FPR)).*

**Explanation: Domain Mismatch Between Natural Images and Neuroimaging Drives Failures**

Why do these widely used methods perform so poorly in the neuroimaging context? We hypothesized that these failures stem from a fundamental domain mismatch: methods developed, tuned, or validated primarily on natural images may rely on assumptions or heuristics that do not hold for neuroimaging data (cf. SM-D5). Natural images typically contain well-defined objects with sharp edges, compositional hierarchies, and specific texture statistics, whereas brain MRIs are volumetric, possess strong long-range spatial correlations, and often involve subtle, diffuse, or non-geometric features of interest (cf. Marek et al., 2022; Mechelli et al., 2005; Schulz et al., 2020).

To test this hypothesis, we directly compared the performance of the *same* XAI method implementations on our neuroimaging benchmark tasks against their performance on a standard natural image benchmark dataset (ImageNet). The results revealed a remarkable divergence (Fig. 4, ST-11). Methods that performed poorly on our neuroimaging tasks, namely LRP and GradCAM, achieved high RMA scores on the natural image benchmark, consistent with their perceived effectiveness in that domain. Conversely, SmoothGrad, which proved most reliable in our neuroimaging framework, exhibited comparatively lower performance on the natural image benchmark. This inverse performance ranking suggests that the design principles or implicit biases of methods like LRP and GradCAM are indeed tailored to natural image characteristics and fail to generalize effectively to the distinct properties of 3D brain MRI data. This underscores the critical importance of domain-specific validation and the potential pitfalls of naively transferring XAI tools across disparate data modalities.

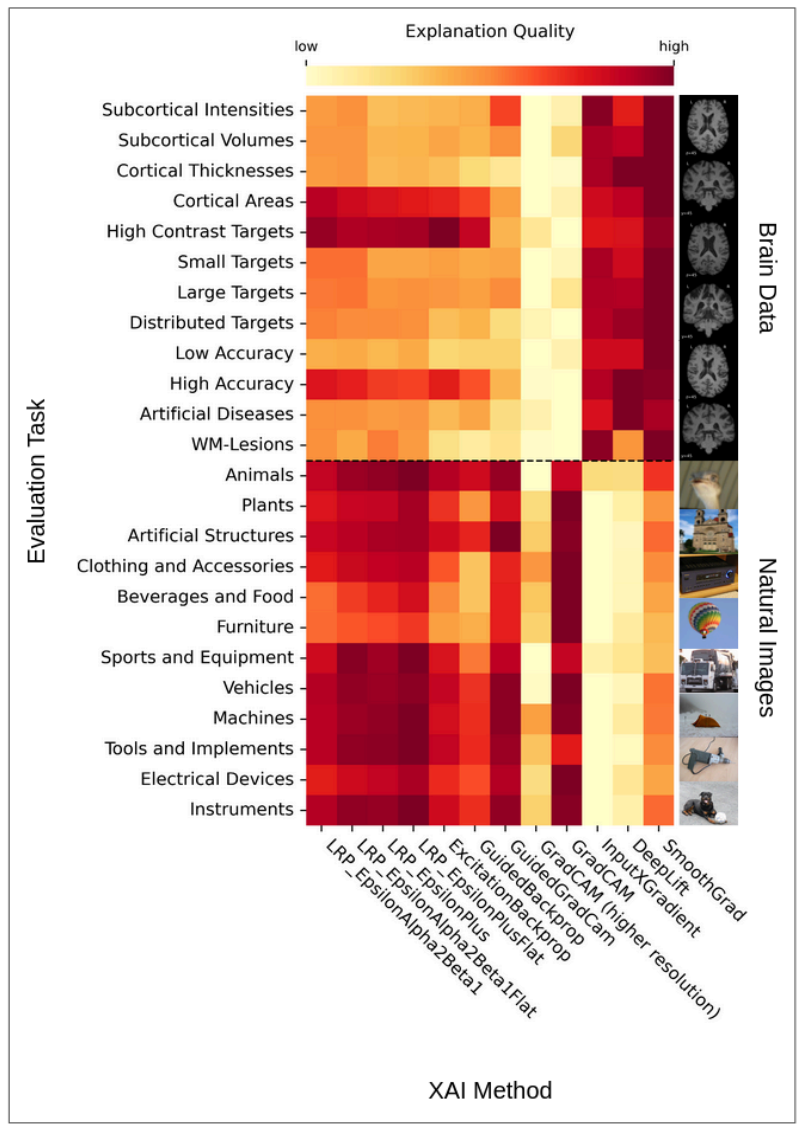

*Figure 4: Domain-specific evaluation is crucial: XAI method performance diverges between neuroimaging and natural image domains. Systematic benchmarking using the same methods, the same domain-adapted model architecture (3D ResNet for brain, 2D for images), and metric for explanation quality (Relevance Mass Accuracy - RMA) reveals contrasting performance patterns. Methods performing well on natural images (e.g., LRP, GradCAM) show poor performance on our neuroimaging tasks. Conversely, SmoothGrad, the top performer on neuroimaging, shows weaker performance on natural images (quantitative results in ST-11, row-wise min-max scaled RMAs underlying Fig. 4 in ST-5). This performance inversion highlights a domain mismatch, indicating that methods optimized for one domain may not be reliable for the other, necessitating domain-specific validation for trustworthy explanations (details in SM-E).*

**Solution: SmoothGrad as a Validated Alternative for Interpretation**

In contrast to the failures of LRP and GradCAM, our validation framework identified gradient-based methods, particularly SmoothGrad (Smilkov et al., 2017), as a reliable (Fig. 5) approach for generating trustworthy explanations in structural neuroimaging. SmoothGrad introduces noise to the input multiple times and averages the resulting gradients, which smoothes the explanation map and reduces noise inherent in raw saliency methods.

Quantitatively, SmoothGrad consistently achieved high RMA (Fig. 1c, ST-2) and TPR (Fig. 3b, ST-6) across the spectrum of ground-truth validation tasks, from localized IDPs to distributed lesions, while maintaining a low FPR (Fig. 3a, ST-7). For the complex brain age biomarker, SmoothGrad explanations showed high overlap (Fig. 1c, ST-2) with literature-derived anatomical patterns of aging, supporting their biological plausibility. Qualitatively, SmoothGrad explanations accurately highlighted the ground truth anatomical regions (Fig. 5, further examples in SF-9). For localized IDP tasks (e.g., putamen intensity, caudate volume, gyrus rectus area), explanations were tightly focused on the target structure. For the clinically relevant lesion prediction task, SmoothGrad best identified the location of subject-specific, distributed lesion patterns (Fig. 1c, ST-2).

This robust performance held across tasks targeting features of different types (intensity, volume, thickness, area) and varying sizes, and across models with different levels of predictive accuracy (Fig. 1c; ST-2; details in SM-F2). While inherent gradient noise requires appropriate post-processing (smoothing, thresholding - SM-F3) for clarity, and the precision of *single-subject delineation* for highly complex patterns like lesions may be less sharp than for simple targets (Fig. 5), the overall localization accuracy remains consistently high. The success of this relatively simple method suggests that approaches making fewer assumptions about data structure or feature hierarchies may be inherently more robust to domain shifts.

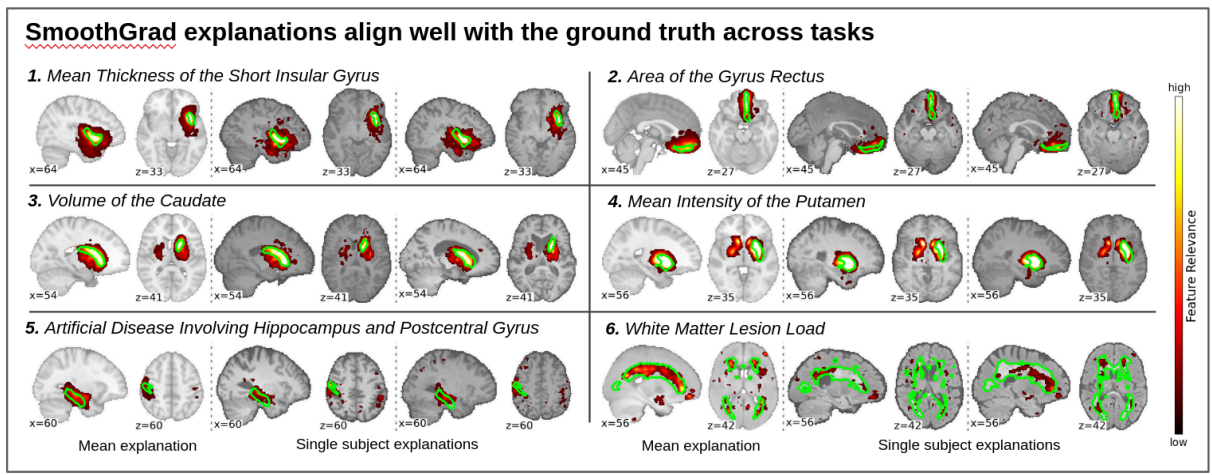

*Figure 5: SmoothGrad explanations faithfully localize ground-truth anatomical features across diverse neuroimaging tasks within the validation framework. Examples show mean explanation maps (first subfigure per task) and representative single-subject explanations (second and third subfigure per task) aligned with ground truth regions (green outlines). Tasks shown: 1. Mean Thickness of Short Insular Gyrus, 2. Area of Gyrus Rectus, 3. Volume of Caudate, 4. Mean Intensity of Putamen, 5. Artificial Disease (Hippocampus + Postcentral Gyrus), 6. White Matter Lesion Load (clinically relevant, subject-specific distributed pattern). High spatial overlap is observed across simple localized features and complex, clinically relevant distributed patterns, validating SmoothGrad's reliability for neuroimaging XAI (quantitative metrics in Fig. 1c, ST-2; further qualitative examples in SF-9).*

**Discussion**

The application of AI in clinical neuroimaging and neuroscience research requires rigorous validation of the tools used to interpret deep learning models. Our study provides the first large-scale, systematic comparison of common XAI methods using a novel validation framework tailored to the unique challenges of structural neuroimaging data. The central finding is concerning: two of the most commonly used XAI methods in the neuroimaging literature, GradCAM in its standard form and LRP with default natural-image rules (survey on method usage in SM-G), exhibit critical and widespread failures - poor localization and artifact generation, respectively - when subjected to scrutiny against verifiable ground truth. This discovery casts doubt on the reliability of interpretations drawn from potentially numerous prior studies (cf. SM-G) that employed these methods without domain-specific validation and highlights a need for methodological correction within the field.

A primary contribution of this work is the development of the validation framework itself. The IDP correction procedure addresses a central challenge in XAI validation: disentangling model failures from explanation method failures. Without correction, raw IDPs exhibit strong brain-wide correlations that permit models to achieve high performance through proxy features, e.g., predicting hippocampal volume indirectly via ventricular size rather than learning to segment the hippocampus itself. In such scenarios, XAI methods face an interpretation ambiguity: explanations highlighting non-target regions could reflect either (1) faithful attribution of the model's reliance on proxy features, or (2) artifactual misattribution by the explanation method. This ambiguity renders objective validation impossible, as both "correct" and "incorrect" explanations become defensible. The cIDP correction resolves this ambiguity by ensuring that predictive performance depends solely on the target anatomical region. As a result, any attribution outside the target can be confidently interpreted as a failure of the XAI method rather than as the model relying on proxy features, which enabled the clear identification of attribution failures in LRP and GradCAM.

By establishing verifiable ground truth across a spectrum of complexity - from precisely localized anatomical targets (corrected IDPs) to real-world clinical features (lesions) and literature-based patterns (brain age) - while preserving the integrity of the input neuroimaging data, this framework provides a much-needed, objective methodology for evaluating XAI reliability. Its structure, moving from controlled simplicity to clinical complexity, was essential for definitively identifying the systemic nature of the failures in GradCAM and LRP, and for building confidence in the reliability of SmoothGrad. We propose this ground-truth target-based validation approach as a standard for future evaluations of XAI methods in neuroimaging and potentially other specialized medical imaging domains.

The marked divergence in method performance between our neuroimaging benchmark and standard natural image datasets (Fig. 4) provides evidence for domain mismatch as the cause of these failures. Methods like GradCAM, relying heavily on final convolutional layer activations (Selvaraju et al., 2017), may falter when relevant information in neuroimaging models is represented differently, perhaps in earlier layers or through non-hierarchical spatial relationships (cf. SM-D5). LRP, with its various propagation rules often selected for visual appeal and object localization performance on natural images (Bach et al., 2015; Kohlbrenner et al., 2020), may require adaptation for the low contrast, often highly distributed region-of-interest patterns in brain MRI, where they appear prone to latching onto high contrast transitions (e.g., ventricles, brain stem), generating artifacts unrelated to true feature importance (Fig. 3a, SM-D2). This finding highlights that AI and XAI tools cannot be assumed to generalize reliably across fundamentally different data domains.

Our results offer practical guidance for the field. Researchers and clinicians should exercise caution when using GradCAM for generating spatial explanations in 3D neuroimaging due to its demonstrated inability to reliably localize relevant features. LRP in its current off-the-shelf configuration should be considered provisional until neuroimaging-adapted rule-sets are available, given its propensity to generate extensive false-positive artifacts that could lead to spurious interpretations. Findings from previous studies relying on these methods, particularly those making strong claims based on the precise spatial location of explanations, may warrant re-evaluation using validated techniques. Our results identify SmoothGrad as a robust and empirically validated alternative. Its consistent performance across our multi-stage framework, including success on challenging subject-specific lesion patterns, suggests its relative simplicity and lack of strong assumptions make it more adaptable to the neuroimaging domain. While not a perfect solution - requiring appropriate post-processing (SM-F3) and acknowledging potential limitations in delineating highly complex patterns at the single-subject level (Fig. 5) - it provides a significantly more reliable starting point for generating trustworthy explanations than the currently prevalent methods.

While our analysis identified several gradient-based methods as reliable (consistent with Wang et al. (2023); Sixt et al. (2019), Adebayo et al. (2018)), our recommendation of SmoothGrad over alternatives like IxG and DeepLift reflects interpretational and practical considerations specific to neuroimaging applications. IxG quantifies feature contributions relative to a zero-input baseline, but this reference point is problematic in neuroimaging contexts where zero-intensity voxels represent non-brain tissue or acquisition artifacts rather than meaningful counterfactuals. Similarly, while DeepLift offers an advantage through its use of reference baseline distributions, practical implementation faces challenges in neuroimaging: large, representative background distributions are computationally infeasible for high-dimensional brain data, zero backgrounds reduce to the questionable IxG case, and mean-intensity backgrounds represent ad-hoc choices that lack principled justification. SmoothGrad circumvents these baseline dependencies through its noise-averaging approach, requiring only the assumption that small perturbations around the input approximate the local gradient manifold - a more defensible assumption for continuous neuroimaging data than arbitrary reference baseline selection.

Our findings do not indicate fundamental flaws in the LRP framework itself but rather highlight the need for domain-specific adaptation of some XAI methods for neuroimaging applications. The strong performance of Input × Gradient - which represents the most basic LRP rule (LRP-0) - suggests that the core relevance propagation principle is applicable, but that the composite LRP rule sets optimized for natural images may be inappropriate for brain MRI data. We outline three plausible mechanisms: (i) edge-biased $\alpha=2$-$\beta=1$ / z+ rules that pull relevance toward high-contrast CSF–tissue boundaries such as ventricles and brain stem; (ii) the property of $\alpha=2$-$\beta=1$ / z+ to downweight or discard inhibitory effects when propagating relevance, which might benefit object localization in image classification, but lead to flawed attributions in regression tasks; (iii) intensity-outlier magnification whereby extreme z-scored values amplify back-propagated relevance in those same structures. Disentangling these factors and designing neuroimaging-specific rule-sets that avoid them will require systematic ablation studies. This suggests a path forward: developing neuroimaging-specific LRP rule configurations, adapted canonization procedures for medical imaging models, and systematic parameter optimization for the unique statistical properties of brain data. More broadly, our results underscore that some explainability tools require deliberate adaptation and validation for new domains rather than wholesale transfer from computer vision benchmarks.

Even perfectly validated AI explanations do not necessarily reflect the underlying biological processes driving clinical predictions. Our experimental design deliberately eliminated confounds to establish methodological ground truth, but real-world neuroimaging datasets contain systematic biases - including scanner effects, demographic imbalances, or subtle data collection artifacts - that can lead models to exploit spurious correlations rather than genuine biological signals (Alexander-Bloch et al., 2016; Chen et al., 2022; Wachinger et al., 2019). Under such conditions, even methodologically sound XAI approaches may produce explanations that accurately reflect what the model learned while misrepresenting the biological relationships of interest. Empirical studies suggest that up to 50% of model explanations in real-world scenarios may reflect such spurious associations (Wang et al., 2023).

Another scenario, where faithful XAI methods may not exclusively highlight the biological relationships of interest, arises in the presence of suppressor variables (Wilming et al., 2022). In such cases, XAI methods may highlight brain regions that help contextualize the main biological effect - potentially leading to misinterpretation of these contextual regions as primary drivers of the effect. These challenges represent not a failure of XAI methodology per se, but rather the broader, intrinsic problem of shortcut learning, confound sensitivity, and suppressor variables in machine learning applications.

This work represents a crucial step towards building trust in the application of deep learning in neuroimaging. By critically evaluating interpretation methods and providing a validated approach, we enable researchers to move beyond simple prediction towards more reliable insights into the features driving model decisions, facilitating safer clinical translation and more robust neuroscientific discovery. However, several limitations should be acknowledged. The considerable computational expense inherent to this study constrained its experimental breadth. A comprehensive evaluation across a wider range of architectures (e.g., Transformers) was beyond the current scope. Similarly, resource constraints limited our analysis primarily to T1-weighted and T2 FLAIR MRI and prevented a more exhaustive exploration with a greater variety of IDPs. Further work is needed to validate these findings across other modalities (fMRI, DWI), architectures (e.g., Transformers (Siegel et al., 2025)), and diverse clinical datasets. The corrected IDPs, while providing localized ground truth, represent abstract features whose direct clinical correlates require further investigation. Furthermore, the brain age validation remains a plausibility check against literature, not absolute ground truth. Future research should aim to extend this validation framework, potentially incorporating causal concepts, longitudinal data, and more sophisticated ground-truth paradigms, and strive to develop novel XAI methods specifically designed for the unique characteristics of neuroimaging data. In this manuscript, we identified failure modes of common XAI methods in their application to neuroimaging data, provided evidence that these failures stem from a domain mismatch between natural and brain images, and therefore recommend minimal-assumption gradient-based methods for trustworthy application of XAI in neuroimaging.

## Methods

### Dataset and Preprocessing

Neuroimaging data were obtained from the UK Biobank resource (Application 33073), selected for its large scale and standardized acquisition protocols, comprising T1-weighted (T1w) and T2-weighted FLAIR structural MRI scans from 45,760 participants after quality control. Data preprocessing involved standard steps including bias field correction, brain extraction, and linear registration to MNI152 standard space to produce analysis-ready images at 1 mm isotropic resolution, ensuring comparability across subjects. Full cohort details, acquisition parameters, and preprocessing steps are provided in SM-A2.

### XAI Validation Framework

We developed a validation framework designed to systematically evaluate XAI methods against verifiable ground truth across tasks of increasing complexity, crucially without modifying the input MRI data to maintain realism and preserve the natural statistics of the data. The progression from simple localized features to complex clinical patterns allows for a nuanced assessment of method capabilities and failure modes (Framework rationale in SM-A1).

**Stage 1: Localized Anatomical Features (Corrected IDPs)** The framework's first stage establishes ground truth for tasks with a single, verifiably localized predictive signal. We began with standard Imaging-Derived Phenotypes (Alfaro-Almagro et al., 2018) - quantitative measures of regional anatomy like volume or thickness. However, raw IDPs are unsuitable for ground truth validation because they exhibit widespread correlations across the brain, driven by global factors such as head size, age-related atrophy, or MRI scanner effects. A model predicting a raw IDP could thus rely on features far outside the target anatomical region.
To address this, we developed a correction method to produce corrected IDPs (cIDPs) whose variance is almost exclusively driven by local anatomy. For each target IDP, we first compiled a large set of related anatomical measures from which the target's family was excluded (e.g., using all other subcortical volumes to correct the hippocampal volume). We then applied Principal Component Analysis (PCA) to this set to extract components representing the major axes of shared, global variance. The raw IDP was regressed against these components, and the residual from this regression became the cIDP. This procedure removes the confounding global variance, isolating a signal specific to the target structure. The number of components removed was optimized for each IDP to maximize this anatomical localization, guided by spatial correlation maps (see SM-A3 for full methodology).

This process yields a prediction target (the cIDP) for which the corresponding anatomical region's mask - e.g. provided by the Destrieux brain atlas (Destrieux et al., 2010) - serves as the ground truth for explanation evaluation. We validated this localization using a masking experiment: when the target anatomical region was computationally removed from the input images, a deep learning model's ability to predict the cIDP collapsed ($R^2 \approx 0$). This result confirms that the predictive signal is causally dependent on the target region, validating its use as a ground truth standard for XAI evaluation (full validation results in our Companion Manuscript, Table A2).

Effects of the cIPD procedure on the causal structure of the prediction problem are described in SM-A3.

**Stage 2: Controlled Distributed Patterns ("Artificial Diseases")** To assess whether XAI methods are sensitive to distributed predictive signals in the brain, we created two "artificial diseases" - synthetic binary classification targets derived by combining cIDPs from distinct cortical and subcortical regions. Each disease label was assigned based on subjects exhibiting high values for one cIDP and low values for another (above the 60th and at or below the 40th percentile, respectively), with mid-percentile cases excluded to sharpen class boundaries. This process yielded imbalanced datasets (~1:3 patient-to-control ratio), but despite this imbalance, the models achieved high classification performance (accuracy > 0.80), confirming that the synthetic labels carried learnable information. Ground truth masks for evaluation explanations were constructed by combining the anatomical regions tied to each cIDP. Full methodological details, including the specific cIDPs used to define each artificial disease, are provided in SM-A4.

**Stage 3: Clinically Relevant Distributed Patterns (Lesions)** To evaluate XAI methods in a real-world clinical context, we trained a model to predict individual white matter hyperintensity (WMH) lesion load - a common and clinically significant imaging - from T2 FLAIR MRI scans. WMHs present as distributed, heterogeneous patterns that vary in location and extent across individuals, offering a realistic and pathologically grounded testbed for XAI methods. The model achieved strong predictive performance ($R^2 = 0.93$). Subject-specific WMH segmentations, derived using the BIANCA tool and provided by the UK Biobank, served as ground truth for evaluating model explanations. (Details in SM-A5).

**Stage 4: Complex Biomarker Plausibility (Brain Age)** We trained models to predict chronological age from T1-weighted structural MRIs (Brain Age Prediction), a complex biomarker where ground truth is not directly localized. Explanation plausibility was therefore assessed by quantitatively comparing the spatial distribution of model explanations against 17 established anatomical markers of aging identified in a large-scale literature meta-analysis by (Walhovd et al., 2011), testing alignment with known biological processes (Thomas et al., 2023; Wang et al., 2023). Anatomical brain regions - defined by the Destrieux atlas (Destrieux et al., 2010) and Freesurfer ASEG subsegmentations (Fischl et al., 2002) - were ranked by a relevance score based on the 99th percentile of explanation values within each region's anatomical mask. Alignment was then evaluated by measuring the overlap between each participant's top-ranked regions and the literature-based aging markers (Details in SM-A6).

**Deep Learning Models** For the main text results, we used a standard 3D ResNet-50 architecture (Hara et al., 2018), chosen for its common use and strong performance in medical imaging (replication on different architecture in SM-F4). The model was adapted for regression when predicting continuous targets (cIDPs, lesion load, brain age), and for binary classification in the artificial disease task. For regression, models were trained using Mean Squared Error loss; for classification, Binary Cross-Entropy loss was used. All models were optimized using Adam (Kingma & Ba, 2014), the de facto standard optimizer in deep learning for both regression and classification tasks. Full architecture specifications, training parameters ensuring convergence, data splits, and model performance metrics demonstrating adequate learning for all tasks are provided in SM-B1 and SM-B2.

**Explainable AI (XAI) Methods Implementation** We evaluated a comprehensive suite of XAI methods, selected to represent the major conceptual classes (gradient-based, relevance-based, reference-based, CAM-based) and include those most commonly applied in the neuroimaging literature (see SM-G). **Gradient-based**: SmoothGrad (Smilkov et al., 2017), Input × Gradient (Shrikumar et al., 2017), Guided Backpropagation (Springenberg et al., 2014), Excitation Backprop (J. Zhang et al., 2018) **Relevance-based**: Layer-wise Relevance Propagation (LRP) (Bach et al., 2015), including multiple rule variants (e.g., LRP-EpsilonAlpha2Beta1, LRP-EpsilonPlus; see SM-B3 for details). **Reference-based**: DeepLift (Shrikumar et al., 2017), using the population mean T1w image as baseline. **CAM-based**: GradCAM and Guided GradCAM (Selvaraju et al., 2017), using activations from the last convolutional layer (analysis of other layers in SM-B3).

Methods were implemented using established libraries where possible, with parameters chosen based on common practices or preliminary evaluations. Implementation details, library versions, specific parameters for all methods, justifications, and necessary post-processing steps (e.g., smoothing/thresholding for SmoothGrad, with sensitivity analyses in SM-F3) are provided in Supplementary Analyses SM-B3 and SM-F3.

**Explanation Evaluation Metrics**

Explanation quality was primarily assessed using established metrics (Arras et al., 2022), chosen to capture complementary aspects of explanation fidelity: Relevance Mass Accuracy (RMA), measuring the proportion of absolute explanation signal correctly localized within the ground truth mask; True Positive Rate (TPR), measuring the percentage of cases where the ground-truth region was successfully identified (among the three most salient brain regions) by the explanation; and False Positive Rate (FPR), measuring the percentage of cases where explanations assigned high relevance to regions verifiably unrelated to the ground truth target. Full details on all evaluation metrics, including formal definitions, are provided in Supplementary Material SM-B4. Explanation postprocessing steps and sensitivity analyses regarding metric dependence on explanation map thresholding and are provided in SM-F3.

**Natural Image Benchmark Comparison**

To explicitly test the domain mismatch hypothesis – that XAI method performance differs between imaging domains - we compared method performance (using RMA) on our 3D neuroimaging tasks to a 2D natural image benchmark, using a subset of ImageNet with object segmentation masks serving as proxies for ground truth explanations. We used the 2D counterpart of our 3D ResNet-50 architecture to ensure consistency across domains. This setup enables a direct assessment of how domain differences impact XAI performance. Details of the natural image benchmark setup and comparative results are provided in Supplementary Material SM-E.

**Statistical Analysis and Visualization**

Quantitative metrics were computed per subject and averaged across the test set to assess typical performance. Group-level explanations were generated by averaging individual maps to visualize common patterns. Visualizations used standard neuroimaging libraries (e.g., Nilearn). Standard statistical tests were employed for comparisons where appropriate. Comprehensive quantitative results are provided in SM-C, SM-D, and SM-E.

**Code and Data Availability**

The code used for preprocessing, model training, XAI method implementation, and evaluation is available at [GitHub Repository Link]. Processed data and results necessary to reproduce the main findings are available at [Data Repository Link]. Raw UK Biobank data are available upon application via the UK Biobank Access Management System.

**Acknowledgments**

We thank Grégoire Montavon, and Wojciech Samek for constructive discussions on LRP implementation. We thank the UKBB participants for their voluntary commitment and the UKBB team for their work in collecting, processing, and disseminating these data for analysis. Research was conducted using the UKBB resource under project-ID 33073. Computation has been performed on the HPC Cluster of the Charité – Universitätsmedizin Berlin. The project was funded by Deutsche Forschungsgemeinschaft (DFG; 414984028, 389563835, 402170461, 459422098, 442075332 to KR), the Hertie Foundation, the Brain & Behavior Research Foundation (NARSAD young investigator grant to KR), a DMSG research award (KR), and by the NIH (5R01AG080821, 1R01AG085571, and 5R01AG083865 to MH). The funders had no role in study design, data collection and analysis, decision to publish, or preparation of the manuscript. This manuscript was edited for style and grammar using LLMs.

**Supplementary Online Material**

SM-A: Validation Framework: Methodology and Ground Truth Establishment
*   (A1) Framework Overview & Rationale
*   (A2) Dataset Details
*   (A3) Stage 1: Corrected IDP Generation and Validation
*   (A4) Stage 2: Controlled Distributed Patterns ("Artificial Diseases")
*   (A5) Stage 3: Clinically Relevant Distributed Patterns (Lesions)
*   (A6) Stage 4: Complex Biomarker Plausibility (Brain Age)

SM-B: Deep Learning Model and XAI Implementation Details
*   (B1) Model Architecture
*   (B2) Model Training
*   (B3) XAI Method Implementation
*   (B4) Evaluation Metrics

SM-C: Comprehensive Quantitative Benchmark Results
*   (Tables and figures with RMA, TPR, FPR, and plausibility scores)

SM-D: Qualitative Evaluation and Failure Mode Analysis
*   (D1) Conflicting Explanations Deep Dive
*   (D2) LRP Artifact Showcase
*   (D3) GradCAM Localization Failure Showcase
*   (D4) SmoothGrad Qualitative Performance
*   (D5) Extended Discussion of Domain Mismatch Effects on LRP and GradCAM

SM-E: Domain Mismatch Investigation
*   (E1) Natural Image Benchmark Setup
*   (E2) Cross-Domain Performance Comparison

SM-F: Method Sensitivity, Robustness, and Post-Processing
*   (F1-F5) Analyses of method sensitivity, robustness, and post-processing effects.

SM-G: XAI Method Usage in Neuroimaging Literature
*   (Systematic literature search results)

Supplementary Tables (see additional .csv files):
*   (ST-1) Predictive Performance
*   (ST-2) Quantitative XAI Performance (Including RMA Scores and Aging Marker Overlap)
*   (ST-3) Standard Deviation Across the Population for Scores in ST-2
*   (ST-4) RMA and Aging Marker Overlap Scores (Row-Wise Min-Max Scaled) for Figure 1c
*   (ST-5) RMA Scores (Row-Wise Min-Max Scaled) for Figure 4 (Domain Mismatch)
*   (ST-6) TPR across cIDP Tasks
*   (ST-7) FPR across cIDP Tasks
*   (ST-8) Statistical Testing Results for Comparison of Main XAI Methods
*   (ST-9) Mapping Between ImageNet Classes and Semantic Supercategories
*   (ST-10) Number of Used Segmentation Masks for Each ImageNet Supercategory
*   (ST-11) Full Results of Domain Mismatch Experiment (RMA scores)
*   (ST-12) Standard Deviation Across the Population for Scores in ST-11
*   (ST-13) Target Region Sizes in mm³
*   (ST-14) Quantitative XAI Performance (Including RMA Scores and Aging Marker Overlap) for Best-Performing Thresholds

* (ST-15) ImageNet XAI Performance (RMA Scores) for Best-Performing Thresholds
* (ST-16) Predictive Performance for Alternative Architecture
* (ST-17) Alternative Architecture: Quantitative XAI Performance (Including RMA Scores and Aging Marker Overlap)
* (ST-18) Standard Deviation Across the Population for Scores in ST-17
* (ST-19) Alternative Architecture: RMA and Aging Marker Overlap Scores (Row-Wise Min-Max Scaled) for Supplementary Material Figure SM-F.F1
* (ST-20) Alternative Architecture: Quantitative XAI Performance (Including RMA Scores and Aging Marker Overlap) for Best-Performing Thresholds

Supplementary Figures (see additional .pdf files):
* (SF-1) Extended LRP Showcase (EpsilonAlpha2Beta1)
* (SF-2) Extended LRP Showcase (EpsilonAlpha2Beta1Flat)
* (SF-3) Extended LRP Showcase (EpsilonPlus)
* (SF-4) Extended LRP Showcase (EpsilonPlusFlat)
* (SF-5) Extended GradCAM Showcase (Last Layer Activations)
* (SF-6) Extended GradCAM Showcase (3rd Layer Block Activations)
* (SF-7) Extended GradCAM Showcase (2nd Layer Block Activations)
* (SF-8) Extended GradCAM Showcase (1st Layer Block Activations)
* (SF-9) Extended SmoothGrad Showcase

Companion Manuscript: Generation of Anatomically Localized Imaging-Derived Phenotype Targets for Ground Truth Validation of Explainable AI in Neuroimaging

# References

Adebayo, J., Gilmer, J., Muelly, M., Goodfellow, I., Hardt, M., & Kim, B. (2018). Sanity checks for saliency maps. *Advances in Neural Information Processing Systems*, *31*.

Alexander-Bloch, A., Clasen, L., Stockman, M., Ronan, L., Lalonde, F., Giedd, J., & Raznahan, A. (2016). Subtle in-scanner motion biases automated measurement of brain anatomy from in vivo MRI. *Human Brain Mapping*, *37*(7), 2385–2397.

Alfaro-Almagro, F., Jenkinson, M., Bangerter, N. K., Andersson, J. L. R., Griffanti, L., Douaud, G., Sotiropoulos, S. N., Jbabdi, S., Hernandez-Fernandez, M., Vallee, E., Vidaurre, D., Webster, M., McCarthy, P., Rorden, C., Daducci, A., Alexander, D. C., Zhang, H., Dragonu, I., Matthews, P. M., … Smith, S. M. (2018). Image processing and Quality Control for the first 10,000 brain imaging datasets from UK Biobank. *NeuroImage*, *166*, 400–424.

Arras, L., Osman, A., & Samek, W. (2022). CLEVR-XAI: A benchmark dataset for the ground truth evaluation of neural network explanations. *An International Journal on Information Fusion*, *81*, 14–40.

Bach, S., Binder, A., Montavon, G., Klauschen, F., Müller, K.-R., & Samek, W. (2015). On pixel-wise explanations for non-linear classifier decisions by layer-wise relevance propagation. *PloS One*, *10*(7), e0130140.

Böhle, M., Eitel, F., Weygandt, M., & Ritter, K. (2019). Layer-wise relevance propagation for explaining deep neural network decisions in MRI-based Alzheimer's disease classification. *Frontiers in Aging Neuroscience*, *11*, 194.

Bommer, P. L., Kretschmer, M., Hedström, A., Bareeva, D., & Höhne, M. M.-C. (2024). Finding the right XAI method—A guide for the evaluation and ranking of Explainable AI methods in climate science. *Artificial Intelligence for the Earth Systems*, *3*(3). https://doi.org/10.1175/aies-d-23-0074.1

Brier, M. R., Li, Z., Ly, M., Karim, H. T., Liang, L., Du, W., McCarthy, J. E., Cross, A. H., Benzinger, T. L. S., Naismith, R. T., & Chahin, S. (2023). "Brain age" predicts disability accumulation in multiple sclerosis. *Annals of Clinical and Translational Neurology*, *10*(6), 990–1001.

Budding, C., Eitel, F., Ritter, K., & Haufe, S. (2021). Evaluating saliency methods on artificial data with different background types. In *arXiv [eess.IV]*. arXiv. http://arxiv.org/abs/2112.04882

Chen, A. A., Beer, J. C., Tustison, N. J., Cook, P. A., Shinohara, R. T., Shou, H., & Alzheimer's Disease Neuroimaging Initiative. (2022). Mitigating site effects in covariance for machine learning in neuroimaging data. *Human Brain Mapping*, *43*(4), 1179–1195.

Cole, J. H., & Franke, K. (2017). Predicting age using neuroimaging: innovative brain ageing biomarkers. *Trends in Neurosciences*, *40*(12), 681–690.

Cole, J. H., Raffel, J., Friede, T., Eshaghi, A., Brownlee, W. J., Chard, D., De Stefano, N., Enzinger, C., Pirpamer, L., Filippi, M., Gasperini, C., Rocca, M. A., Rovira, A., Ruggieri, S., Sastre-Garriga, J., Stromillo, M. L., Uitdehaag, B. M. J., Vrenken, H., Barkhof, F., … MAGNIMS study group. (2020). Longitudinal assessment of multiple sclerosis with the brain-age paradigm: Brain-age paradigm in multiple sclerosis. *Annals of Neurology*, *88*(1), 93–105.

Debette, S., & Markus, H. S. (2010). The clinical importance of white matter hyperintensities on brain magnetic resonance imaging: systematic review and meta-analysis. *BMJ (Clinical Research Ed.)*, *341*(jul26 1), c3666.

Destrieux, C., Fischl, B., Dale, A., & Halgren, E. (2010). Automatic parcellation of human cortical gyri and sulci using standard anatomical nomenclature. *NeuroImage*, *53*(1), 1–15.

Doshi-Velez, F., & Kim, B. (2017). Towards A rigorous science of interpretable machine learning. In *arXiv [stat.ML]*. arXiv. http://arxiv.org/abs/1702.08608

Eitel, F., Soehler, E., Bellmann-Strobl, J., Brandt, A. U., Ruprecht, K., Giess, R. M., Kuchling, J., Asseyer, S., Weygandt, M., Haynes, J.-D., Scheel, M., Paul, F., & Ritter, K. (2019). Uncovering convolutional neural network decisions for diagnosing multiple sclerosis on conventional MRI using layer-wise relevance propagation. *NeuroImage. Clinical*, *24*(102003), 102003.

Fischl, B., Salat, D. H., Busa, E., Albert, M., Dieterich, M., Haselgrove, C., van der Kouwe, A., Killiany, R., Kennedy, D., Klaveness, S., Montillo, A., Makris, N., Rosen, B., & Dale, A. M. (2002). Whole brain segmentation: automated labeling of neuroanatomical structures in the human brain. *Neuron*, *33*(3), 341–355.

Gilpin, L. H., Bau, D., Yuan, B. Z., Bajwa, A., Specter, M., & Kagal, L. (2018, October). Explaining explanations: An overview of interpretability of machine learning. *2018 IEEE 5th International Conference on Data Science and Advanced Analytics (DSAA)*. 2018 IEEE 5th International Conference on Data Science and Advanced Analytics (DSAA), Turin, Italy. https://doi.org/10.1109/dsaa.2018.00018

Habes, M., Erus, G., Toledo, J. B., Zhang, T., Bryan, N., Launer, L. J., Rosseel, Y., Janowitz, D., Doshi, J., Van der Auwera, S., von Sarnowski, B., Hegenscheid, K., Hosten, N., Homuth, G., Völzke, H., Schminke, U., Hoffmann, W., Grabe, H. J., & Davatzikos, C. (2016). White matter hyperintensities and imaging patterns of brain ageing in the general population. *Brain: A Journal of Neurology*, *139*(Pt 4), 1164–1179.

Hahn, T., Ernsting, J., Winter, N. R., Holstein, V., Leenings, R., Beisemann, M., Fisch, L., Sarink, K., Emden, D., Opel, N., Redlich, R., Repple, J., Grotegerd, D., Meinert, S., Hirsch, J. G., Niendorf, T., Endemann, B., Bamberg, F., Kröncke, T., … Berger, K. (2022). An uncertainty-aware, shareable, and transparent neural network architecture for brain-age modeling. *Science Advances*, *8*(1), eabg9471.

Hara, K., Kataoka, H., & Satoh, Y. (2018). Can spatiotemporal 3d cnns retrace the history of 2d cnns and imagenet? *Proceedings of the IEEE Conference on Computer Vision and Pattern Recognition*, 6546–6555.

Hofmann, S. M., Beyer, F., Lapuschkin, S., Goltermann, O., Loeffler, M., Müller, K.-R., Villringer, A., Samek, W., & Witte, A. V. (2022). Towards the interpretability of deep learning models for multi-modal neuroimaging: Finding structural changes of the ageing brain. *NeuroImage*, *261*(119504), 119504.

Holzinger, A., Langs, G., Denk, H., Zatloukal, K., & Müller, H. (2019). Causability and explainability of artificial intelligence in medicine. *Wiley Interdisciplinary Reviews. Data Mining and Knowledge Discovery*, *9*(4), e1312.

Isensee, F., Jaeger, P. F., Kohl, S. A. A., Petersen, J., & Maier-Hein, K. H. (2021). nnU-Net: a self-configuring method for deep learning-based biomedical image segmentation. *Nature Methods*, *18*(2), 203–211.

Kaufmann, T., van der Meer, D., Doan, N. T., Schwarz, E., Lund, M. J., Agartz, I., Alnæs, D., Barch, D. M., Baur-Streubel, R., Bertolino, A., Bettella, F., Beyer, M. K., Bøen, E., Borgwardt, S., Brandt, C. L., Buitelaar, J., Celius, E. G., Cervenka, S., Conzelmann, A., … Westlye, L. T. (2019). Common brain disorders are associated with heritable patterns of apparent aging of the brain. *Nature Neuroscience*, *22*(10), 1617–1623.

Kelly, C. J., Karthikesalingam, A., Suleyman, M., Corrado, G., & King, D. (2019). Key challenges for delivering clinical impact with artificial intelligence. *BMC Medicine*, *17*(1), 195.

Kindermans, P.-J., Hooker, S., Adebayo, J., Alber, M., Schütt, K. T., Dähne, S., Erhan, D., & Kim, B. (2019). The (Un)reliability of saliency methods. In *Explainable AI: Interpreting, Explaining and Visualizing Deep Learning* (pp. 267–280). Springer International Publishing.

Kingma, D. P., & Ba, J. (2014). Adam: A method for stochastic optimization. In *arXiv [cs.LG]*. arXiv. http://arxiv.org/abs/1412.6980

Kohlbrenner, M., Bauer, A., Nakajima, S., Binder, A., Samek, W., & Lapuschkin, S. (2020, July). Towards best practice in explaining neural network decisions with LRP. *2020 International Joint Conference on Neural Networks (IJCNN)*. 2020 International Joint Conference on Neural Networks (IJCNN), Glasgow, United Kingdom. https://doi.org/10.1109/ijcnn48605.2020.9206975

Lapuschkin, S., Wäldchen, S., Binder, A., Montavon, G., Samek, W., & Müller, K.-R. (2019). Unmasking Clever Hans predictors and assessing what machines really learn. *Nature Communications*, *10*(1), 1096.

Litjens, G., Kooi, T., Bejnordi, B. E., Setio, A. A. A., Ciompi, F., Ghafoorian, M., van der Laak, J. A. W. M., van Ginneken, B., & Sánchez, C. I. (2017). A survey on deep learning in medical image analysis. *Medical Image Analysis*, *42*, 60–88.

Marek, S., Tervo-Clemmens, B., Calabro, F. J., Montez, D. F., Kay, B. P., Hatoum, A. S., Donohue, M. R., Foran, W., Miller, R. L., Hendrickson, T. J., Malone, S. M., Kandala, S., Feczko, E., Miranda-Dominguez, O., Graham, A. M., Earl, E. A., Perrone, A. J., Cordova, M., Doyle, O., … Dosenbach, N. U. F. (2022). Reproducible brain-wide association studies require thousands of individuals. *Nature*, *603*(7902), 654–660.

Mechelli, A., Friston, K. J., Frackowiak, R. S., & Price, C. J. (2005). Structural covariance in the human cortex. *The Journal of Neuroscience: The Official Journal of the Society for Neuroscience*, *25*(36), 8303–8310.

Montavon, G., Samek, W., & Müller, K.-R. (2018). Methods for interpreting and understanding deep neural networks. *Digital Signal Processing*, *73*, 1–15.

Muehlematter, U. J., Daniore, P., & Vokinger, K. N. (2021). Approval of artificial intelligence and machine learning-based medical devices in the USA and Europe (2015-20): a comparative analysis. *The Lancet. Digital Health*, *3*(3), e195–e203.

Nazir, S., Dickson, D. M., & Akram, M. U. (2023). Survey of explainable artificial intelligence techniques for biomedical imaging with deep neural networks. *Computers in Biology and Medicine*, *156*(106668), 106668.

Oliveira, M., Wilming, R., Clark, B., Budding, C., Eitel, F., Ritter, K., & Haufe, S. (2024). Benchmarking the influence of pre-training on explanation performance in MR image classification. *Frontiers in Artificial Intelligence*, *7*, 1330919.

Pahde, F., Yolcu, G. Ü., Binder, A., Samek, W., & Lapuschkin, S. (2022). Optimizing explanations by network canonization and hyperparameter search. In *arXiv [cs.CV]*. arXiv. http://arxiv.org/abs/2211.17174

Rudin, C. (2019). Stop explaining black box machine learning models for high stakes decisions and use interpretable models instead. *Nature Machine Intelligence*, *1*(5), 206–215.

Schulz, M.-A., Yeo, B. T. T., Vogelstein, J. T., Mourao-Miranda, J., Kather, J. N., Kording, K., Richards, B., & Bzdok, D. (2020). Different scaling of linear models and deep learning in UKBiobank brain images versus machine-learning datasets. *Nature Communications*, *11*(1), 4238.

Selvaraju, R. R., Cogswell, M., Das, A., Vedantam, R., Parikh, D., & Batra, D. (2017, October). Grad-CAM: Visual explanations from deep networks via gradient-based localization. *2017 IEEE International Conference on Computer Vision (ICCV)*. 2017 IEEE International Conference on Computer Vision (ICCV), Venice. https://doi.org/10.1109/iccv.2017.74

Shrikumar, A., Greenside, P., & Kundaje, A. (2017). Learning important features through propagating activation differences. In *arXiv [cs.CV]*. arXiv. http://arxiv.org/abs/1704.02685

Siegel, N. T., Kainmueller, D., Deniz, F., Ritter, K., & Schulz, M.-A. (2025). Do transformers and CNNs learn different concepts of brain age? *Human Brain Mapping*, *46*(8). https://doi.org/10.1002/hbm.70243

Sixt, L., Granz, M., & Landgraf, T. (2019). When explanations lie: Why many modified BP attributions fail. *International Conference on Machine Learning*, *119*, 9046–9057.

Smilkov, D., Thorat, N., Kim, B., Viégas, F., & Wattenberg, M. (2017). SmoothGrad: removing noise by adding noise. In *arXiv [cs.LG]*. arXiv. http://arxiv.org/abs/1706.03825

Springenberg, J. T., Dosovitskiy, A., Brox, T., & Riedmiller, M. (2014). Striving for simplicity: The all convolutional net. In *arXiv [cs.LG]*. arXiv. http://arxiv.org/abs/1412.6806

Thomas, A. W., Ré, C., & Poldrack, R. A. (2023). Benchmarking explanation methods for mental state decoding with deep learning models. *NeuroImage*, *273*(120109), 120109.

Tjoa, E., & Guan, C. (2021). A survey on explainable artificial intelligence (XAI): Toward medical XAI. *IEEE Transactions on Neural Networks and Learning Systems*, *32*(11), 4793–4813.

van der Velden, B. H. M., Kuijf, H. J., Gilhuijs, K. G. A., & Viergever, M. A. (2022). Explainable artificial intelligence (XAI) in deep learning-based medical image analysis. *Medical Image Analysis*, *79*(102470), 102470.

Wachinger, C., Becker, B. G., Rieckmann, A., & Pölsterl, S. (2019). Quantifying confounding bias in neuroimaging datasets with causal inference. In *Lecture Notes in Computer Science* (pp.

484–492). Springer International Publishing.

Walhovd, K. B., Westlye, L. T., Amlien, I., Espeseth, T., Reinvang, I., Raz, N., Agartz, I., Salat, D. H., Greve, D. N., & Fischl, B. (2011). Consistent neuroanatomical age-related volume differences across multiple samples. *Neurobiology of Aging*, *32*(5), 916–932.

Wang, D., Honnorat, N., Fox, P. T., Ritter, K., Eickhoff, S. B., Seshadri, S., Alzheimer's Disease Neuroimaging Initiative, & Habes, M. (2023). Deep neural network heatmaps capture Alzheimer's disease patterns reported in a large meta-analysis of neuroimaging studies. *NeuroImage*, *269*(119929), 119929.

Wilming, R., Budding, C., Müller, K.-R., & Haufe, S. (2022). Scrutinizing XAI using linear ground-truth data with suppressor variables. *Machine Learning*, *111*(5), 1903–1923.

Yang, M., & Kim, B. (2019). Benchmarking Attribution Methods with relative feature importance. In *arXiv [cs.LG]*. arXiv. http://arxiv.org/abs/1907.09701

Zhang, J., Bargal, S. A., Lin, Z., Brandt, J., Shen, X., & Sclaroff, S. (2018). Top-down neural attention by excitation backprop. *International Journal of Computer Vision*, *126*(10), 1084–1102.

Zhang, Y., Song, J., Gu, S., Jiang, T., Pan, B., Bai, G., & Zhao, L. (2023). Saliency-Bench: A comprehensive benchmark for evaluating visual explanations. In *arXiv [cs.CV]*. arXiv. http://arxiv.org/abs/2310.08537

**Generation of Anatomically Localized Imaging-Derived Phenotype Targets for Ground Truth Validation of Explainable AI in Neuroimaging**

**Abstract**

Explainable artificial intelligence (XAI) methods aim to provide insights into the decision-making processes of deep learning models but require systematic validation. In neuroimaging, this validation is particularly challenging since characterizing "correct" explanations is particularly hard. Here, we propose using imaging-derived phenotypes (IDPs) with known anatomical localization for ground-truth XAI evaluation. We create IDPs with known localization by systematically removing global brain effects through principal component analysis, which results in prediction targets verifiably linked to specific brain regions. We demonstrate the efficacy of this approach across 10 diverse IDPs spanning subcortical intensities, regional volumes, cortical thickness, and surface area measurements. Our results show that deep learning models can successfully learn these corrected targets and that their spatial specificity can be validated by selectively masking the target regions. This approach provides a solid foundation for objective evaluation of XAI methods in neuroimaging by establishing anatomically precise ground truth explanations, offering a promising pathway for advancing the interpretability of machine learning in clinical neuroimaging.

**1. Introduction**

Deep learning has fundamentally transformed neuroimaging analysis, achieving unprecedented accuracy in tasks ranging from anatomical segmentation to disease classification and biomarker prediction (Litjens et al., 2017; Shen et al., 2017). However, the opacity of deep neural networks presents a significant barrier to their clinical adoption (Kelly et al., 2019). While a radiologist can explain their diagnostic reasoning through anatomical landmarks and established disease patterns, deep neural networks provide only numerical predictions without inherent interpretability. Explainable AI (XAI) methods have emerged as a potential solution, promising to reveal the features and patterns that drive neural network predictions (Montavon et al., 2018). These approaches generate spatial attribution maps highlighting brain regions that influenced model decisions. However, the reliability of these explanation methods remains largely unverified, particularly in neuroimaging applications where assessing whether an explanation is correct is particularly hard.

The fundamental challenge in validating XAI methods lies in the absence of ground truth—knowing precisely which brain regions should be highlighted in the explanation (Molnar et al., 2020; Ras et al., 2022). This challenge is particularly acute in neuroimaging and its clinical applications, where the relationship between brain structure and function or pathology involves complex, distributed, and potentially unknown patterns. Current validation approaches, such as synthetic lesion pattern insertion (Budding et al., 2021; Hofmann et al., 2022; Oliveira et al., 2024) or comparison against expert annotations (Arun et al., 2021), often oversimplify the problem or introduce subjective biases.

In this work, we introduce a novel approach using imaging-derived phenotypes (IDPs) as prediction targets with known anatomical localization. IDPs represent specific quantitative measures extracted from brain images, such as regional volumes, cortical thickness, or tissue intensities (Alfaro-Almagro et al., 2018). Theoretically, these measures should be determined primarily by local anatomy—for example, the volume of the hippocampus should depend mainly on the hippocampal structure itself. However, in practice, IDPs exhibit widespread correlations across the brain due to global effects like overall brain size, age-related changes, and shared tissue properties.

The presence of these global effects poses a significant confound for XAI validation. If a model predicts a local IDP, such as hippocampal volume, by primarily relying on a global proxy like overall brain size, an attribution map may correctly highlight widespread, distributed features. However, this leaves the researcher unable to determine whether the model has simply learned a valid, albeit uninteresting, global correlation or if the XAI method has failed to identify the specific, localized anatomical structure. Existing approaches to overcome this ambiguity, such as inserting synthetic lesions, circumvent this issue but introduce a new one: they alter the input images. Modifying the input data fundamentally changes the prediction task to one of detecting artificial patterns, rather than learning from the original neuroanatomy. Consequently, the resulting explanations may not reflect how a model or XAI method would perform on authentic clinical data. Our approach avoids this pitfall by creating a localized prediction target without altering the input brain images, thereby preserving the ecological validity of the validation process, while providing verifiable ground-truth for XAI validation.

Our key contribution is a systematic approach to remove these global effects from IDP targets through principal component analysis, resulting in corrected IDPs that are verifiably linked to specific brain regions. This creates prediction targets with known ground truth location for XAI validation while preserving the complexity of real neuroimaging data. We demonstrate the efficacy of this approach across various anatomical features and validate that the spatial specificity is genuine by showing that models cannot learn these targets when the relevant brain regions are masked.

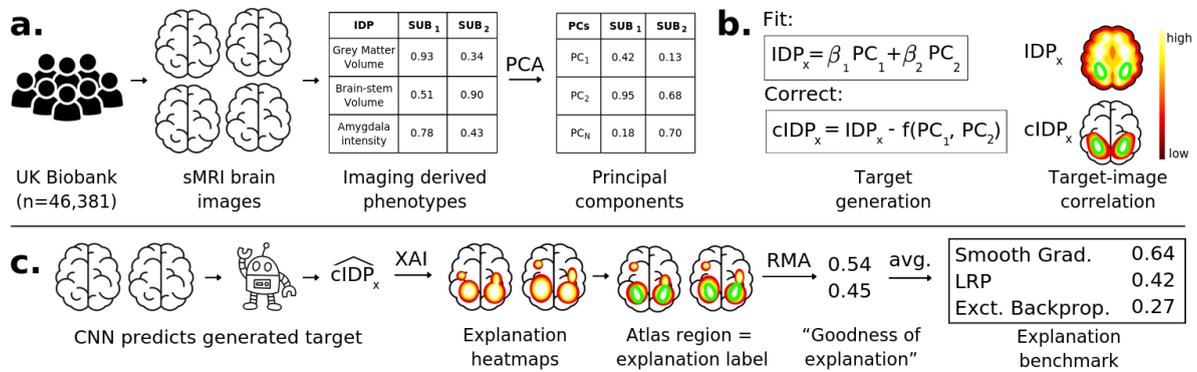

*Figure 1: Overview of the IDP correction and XAI validation pipeline. a) Data flow from UK Biobank participants (n=46,381) through structural MRI acquisition to imaging-derived phenotypes (IDPs) and extraction of principal components. The table shows example IDPs and their principal component representations for two subjects. b) Methodology for generating corrected IDPs (cIDPs) by removing global effects captured in principal components, resulting in localized prediction targets as shown by the target-image correlation maps. c) XAI validation pipeline where a CNN predicts the generated targets, followed by explanation generation and quantitative evaluation of explanation quality using atlas regions as ground truth labels. The table shows performance metrics for three explanation methods, with SmoothGrad achieving the highest relevance mass accuracy (RMA). All numbers in this figure are placeholders.*

The paper is organized as follows: Section 2 describes our methodology for IDP selection, correction, and validation; Section 3 presents the results of our correction procedure and model performance; Section 4 discusses the implications for XAI validation in neuroimaging; and Section 5 concludes with future directions. Detailed technical information for replication is provided in the Appendix.

## 2. Methods

### 2.1 Dataset and IDP Selection

We utilized structural MRI data from the UK Biobank (UKBB) study, which provides standardized, quality-controlled T1-weighted brain scans for a large population cohort (Alfaro-Almagro et al., 2018). From the available UKBB imaging-derived phenotypes, we selected 10 diverse IDP targets representing different anatomical properties and brain regions: subcortical intensities (mean intensity of pallidum and putamen), regional volumes (hippocampus, caudate, brain stem, and lateral ventricle), cortical thickness (postcentral gyrus and insular short gyrus), and cortical surface areas (rectus and orbital gyrus).

The selection criteria included: (1) representation of different brain properties and regions, (2) targets with clear anatomical boundaries defined in standard atlases, and (3) sufficient variability across the population to be learnable by machine learning models. This diverse set of targets allows us to evaluate the generalizability of our approach across different brain structures and measurement types.

## 2.2 IDP Correction Procedure

The central challenge in using raw IDPs as prediction targets for XAI validation is their widespread correlation with global brain characteristics. To address this, we developed a principal component-based correction approach that systematically removes these global effects while preserving the local anatomical signal.

First, we constructed separate IDP sets for correcting cortical and subcortical targets. For subcortical targets, we used 99 IDPs from the UKBB subcortical volumetric segmentation (category 190), while for cortical targets, we used 444 IDPs from the UKBB Destrieux parcellation (category 197). Critically, we removed all IDPs related to the target region from these sets to prevent correcting for the target signal itself. For example, when correcting the "Volume of Hippocampus (left hemisphere)," we removed all hippocampal measures (volume and intensity from both hemispheres) from the correction set.

We then applied principal component analysis (PCA) to these respective IDP sets, capturing the major modes of variation across brain measures. The resulting principal components represent systematic effects that influence multiple brain regions simultaneously, such as overall brain size, global atrophy patterns, or shared tissue properties.

To create corrected targets (cIDPs), we performed linear regression of each raw IDP against an increasing number of principal components (starting from 0 and incrementing by 5) and retained the residuals as the corrected IDP. Each correction level represents a different trade-off between removing global effects and preserving local information. The optimal number of components for correction was determined by visual assessment of the spatial correlation pattern between the corrected IDP and voxel intensities, selecting the level that best localized the signal to the anatomical region of interest.

Correlation between single voxels and *corrected mean intensity of Pallidum*

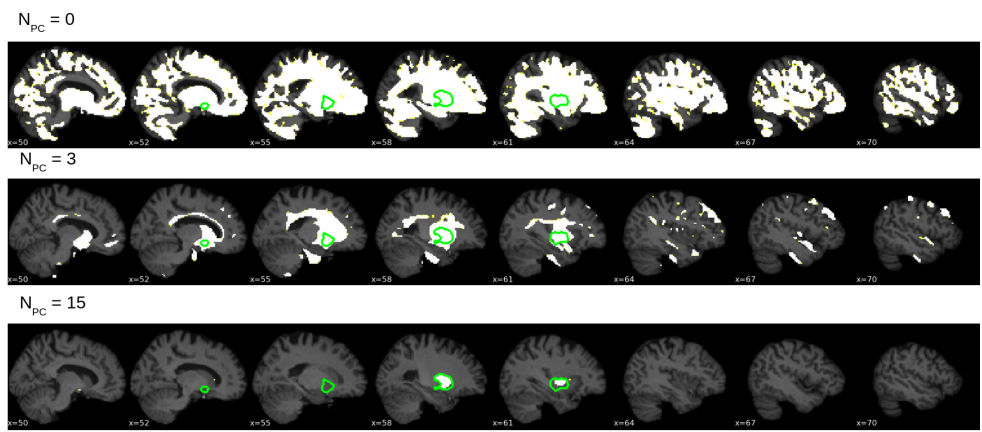

*Figure 2: Progressive anatomical localization of the pallidum signal through principal component correction. Each row shows correlation maps between individual voxels and the mean intensity of the pallidum (right hemisphere) after removing different numbers of principal components (PCs). Top row (N_PC = 0): Without correction, correlations are widespread across the brain. Middle row (N_PC = 3): Partial localization is achieved with minimal PC removal. Bottom row (N_PC = 15): Precise localization to the anatomical region of interest (outlined in green) is achieved, demonstrating successful isolation of local anatomical features from global brain effects. White voxels indicate < 0.05 FWE-corrected significance.*

## 2.3 Validation of Localization

To validate the anatomical specificity of our corrected IDPs, we employed multiple approaches:

First, we computed voxel-wise correlations between each corrected IDP and brain image intensities across the population, visually confirming the spatial localization to the target anatomical region. This analysis revealed how the progressive removal of principal components increasingly focused the correlation pattern on the relevant brain structure. Details of the image processing and statistical analysis procedure are provided in Appendix A.1.

Second, we used a mask-based mass-univariate validation approach to quantitatively assess localization. For each target, we defined an anatomical mask based on the Destrieux atlas (with 20mm dilation to include boundary voxels), and calculated the proportion of significant correlations (alpha = 0.05) falling within this mask compared to whole-brain.

Finally, we conducted a critical test to verify the causal relationship between the target region and the corrected IDP by training deep learning models on images with the target region masked out. Specifically, we used the same ResNet architecture as our main analysis but provided input images where the target anatomical region (dilated by 20mm) was set to zero. If the corrected IDP genuinely represents local anatomical properties, we would expect prediction performance to drop significantly when the relevant region is removed. The detailed methodology for this masking procedure is described in Appendix A.4.

## 2.4 Model Training and Evaluation

To evaluate whether our corrected IDPs remain learnable by deep neural networks, we implemented a standardized deep learning pipeline using 3D ResNets. The pipeline followed the approach used in our brain age prediction work (Schulz et al., 2024; Siegel et al., 2025), with appropriate adaptations for the IDP prediction task.

The dataset of approximately 46,000 subjects was split into training (80%), validation (10%), and test (10%) sets. Models were trained using a ResNet-18 architecture, optimized with the Adam optimizer and a one-cycle learning rate policy. Detailed information about image preprocessing, model architecture, and training parameters is provided in Appendix A.3.

Performance was evaluated using the coefficient of determination ($R^2$) on the test set, providing a measure of how much variance in the corrected IDP could be explained by the model predictions from brain images. The complete results for all IDP targets are presented in Appendix A.5.

## 3. Results

### 3.1 IDP Correction and Anatomical Localization

The application of our PCA-based correction procedure successfully localized the correlation patterns between brain images and IDP targets to their respective anatomical regions. Figure 2 demonstrates this progressive localization for the mean intensity of the right pallidum. Without correction (N_PC = 0), correlations are widespread across the brain, reflecting global effects. With minimal correction (N_PC = 3), a partial localization emerges. The more precise correction (N_PC = 15) shows a highly specific association pattern tightly focused on the pallidum.

Similar localization patterns were achieved for all 10 target IDPs, with the optimal number of principal components varying based on the specific target properties and global correlation structure. The full set of localization results for all targets can be found in the Appendix (Figure A1 and A2), demonstrating the effectiveness of our approach across diverse brain regions and measurement types.

The number of principal components required for optimal correction varied considerably across targets: subcortical intensities required between 20-75 components, volumes between 15-55 components, cortical thickness between 100-325 components, and cortical areas between 145-425 components. This variation aligns with the different degrees to which global effects influence various brain measurements, with cortical surface measures typically requiring extensive correction due to their strong correlations with overall brain morphology (Mechelli et al., 2005).

### 3.2 Model Performance on Corrected IDPs

Despite the removal of global brain effects through our correction procedure, deep learning models were able to successfully learn the corrected IDP targets. The detailed prediction performance ($R^2$) for all targets is presented in the Appendix (Table A2), with values ranging from 0.27 to 0.86. Notably, subcortical volumes and intensities were generally more accurately predicted ($R^2$ = 0.70-0.86) than cortical thickness and area measures ($R^2$ = 0.27-0.52), likely due to the higher variability and noise associated with cortical measurements (Hedges et al., 2022).

The successful prediction of these corrected targets confirms that the localized anatomical information remains learnable by deep neural networks, a critical requirement for their use in XAI validation. The variation in prediction performance across different target types also provides an informative spectrum for evaluating XAI methods under varying conditions of model confidence.

### 3.3 Region Masking Validation

The causal relationship between the target anatomical regions and the corrected IDPs was confirmed through our region masking experiments. When the target region was masked out of the input images, the ResNet models were unable to achieve meaningful prediction performance for any of the corrected IDPs, with $R^2$ values dropping to near zero. For example, the model predicting the corrected mean intensity of the left pallidum achieved an $R^2$ of 0.74 with full brain images but failed to explain any variance when the pallidum was masked out. This was mirrored in our mass-univariate validation results (Table A5)..

These findings provide strong evidence that our correction procedure successfully isolated local anatomical information, as the models specifically rely on the target regions for their predictions rather than exploiting indirect correlations with other brain areas. The full results of these masking experiments are provided in Appendix A.5 (Table A2).

### 3.4 XAI Application

While full evaluation of XAI methods is covered in our companion XAI benchmarking paper, we include here an illustrative example of how the corrected IDPs serve as ground truth for explanation validation. Figure 3 shows SmoothGrad explanations for a model predicting the corrected area of the orbital gyrus. The explanations are consistently localized to the anatomical region of interest, demonstrating alignment with the ground truth target location.

For the explanation evaluation, attribution maps were processed using a standardized pipeline to ensure fair comparisons across methods and subjects. Details of the explanation postprocessing are provided in Appendix A.6.

SmoothGrad explanations for *corrected area of the Orbital gyrus*

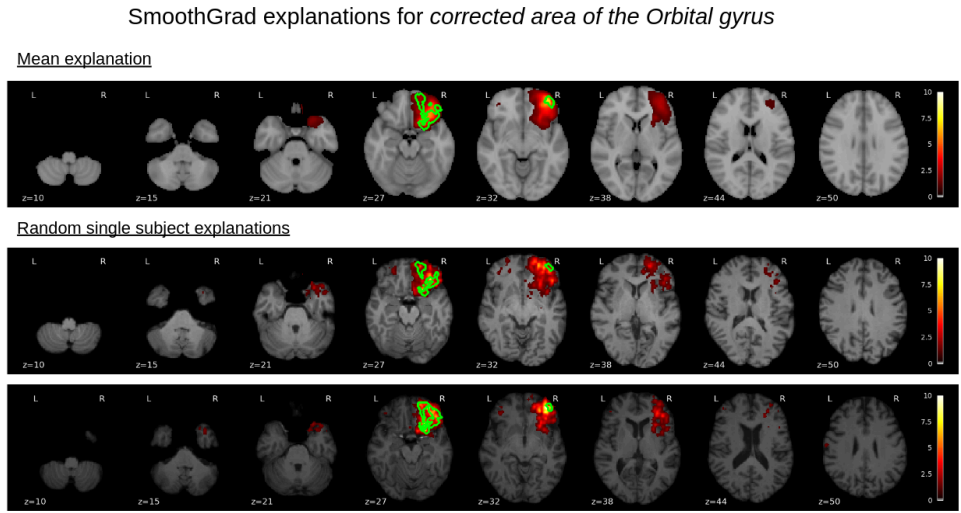

*Figure 3: SmoothGrad explanations for models predicting the corrected area of the orbital gyrus. Top row shows the mean explanation across all test subjects, while the bottom two rows show explanations for individual randomly selected subjects. The green outlines mark the anatomical boundaries of the orbital gyrus according to the Destrieux atlas. Explanation intensities are shown on axial slices (z-coordinates indicated) with a color scale (percentile scaled explanation intensities) where brighter colors indicate stronger influence on the model prediction. Note the consistent localization of explanations to the target region across both average and individual subject maps.*

This qualitative result provides initial support for the utility of our approach in XAI validation, showing that explanations from at least some methods accurately identify the relevant brain regions when models are trained on properly corrected IDP targets. For quantitative evaluation, metrics such as relevance mass accuracy (Arras et al., 2022) can be used to assess how well the explanation aligns with the known anatomical ground truth.

## 4. Discussion

The development of reliable ground truth for validating XAI methods in neuroimaging represents a critical step toward bridging the interpretability gap in clinical applications of deep learning. Our approach using corrected IDPs offers several advantages over existing validation strategies.

First, by creating prediction targets with a verifiable anatomical basis, our approach provides a controlled environment to disentangle the performance of an explanation method from the performance of the underlying model. The validation experiments—both the high prediction accuracy on corrected targets and the performance collapse after region masking—confirm that the models are indeed learning solely from the specified anatomical regions. With this ground truth established, if an XAI method subsequently highlights an incorrect region, the failure can be unambiguously attributed to the explanation method itself, rather than to the model learning from confounding global features or unexpected proxy variables.

Second, our method preserves the ecological validity of the data by using unaltered brain images. Unlike approaches that rely on inserting synthetic lesions or patterns, our framework challenges models and their corresponding explanation methods with the full complexity and variability of real neuroimaging data. This is critical because modifying input images transforms the task into one of detecting artificial signals, which may not be representative of how a model processes subtle, naturally-occurring anatomical variations in a clinical context. By ensuring that the validation setup mirrors the real-world application, we increase the likelihood that findings on XAI performance will generalize beyond the benchmark to actual clinical use cases.

Third, the diverse set of anatomical targets spanning different brain properties and regions enables a comprehensive evaluation of XAI methods across varying conditions. Some targets, like subcortical volumes, provide clear anatomical boundaries and high predictability, establishing a robust baseline for evaluation. Others, such as cortical thickness measurements, represent more challenging scenarios with less distinct delineations and lower signal-to-noise ratios (Hedges et al., 2022). This allows for systematically probing how XAI performance varies with target characteristics like region size, tissue type, and measurement modality (e.g., volume, thickness, area, or intensity).

Furthermore, the framework of corrected IDPs can be extended to model more complex predictive patterns. By combining multiple cIDPs, one could construct targets representing distributed networks of brain regions (a logical AND), testing whether an XAI method can correctly identify all contributing sources. This would be a step toward validating explanations for complex network-based pathologies. Conversely, one could create disjunctive targets (a logical OR), where a phenotype is driven by one of several possible regions in different individuals. Such a scenario would test an XAI method's ability to delineate patient-specific versus general predictive patterns, a critical capability for personalized medicine.

The observed variations in the number of principal components required for optimal correction across different brain measures offer insights into the global correlation structure of brain morphology. Cortical surface measures required substantially more principal components for proper localization compared to subcortical volumes and intensities, suggesting stronger global influences on cortical morphometry. This aligns with known patterns of structural covariance in the brain, where cortical regions show coordinated developmental and aging patterns (Mechelli et al., 2005).

The successful prediction of corrected IDPs by deep learning models, despite the removal of global effects, confirms that these targets retain learnable anatomical information. This finding is crucial, as it demonstrates that our approach does not simply create artificial targets but rather isolates genuine local anatomical signals that can be detected from brain images.

Our region masking validation provides strong evidence that causal relationships between non-target anatomical regions and the corrected IDPs have been effectively removed. The models' failure to predict targets when the relevant regions are masked confirms that our correction procedure has successfully removed indirect correlations with other brain areas, resulting in truly localized targets.

The development of this validation framework addresses a fundamental challenge in the field of explainable AI for neuroimaging. By providing objective ground truth for model explanations, it enables systematic evaluation of different XAI methods and informs the development of more reliable approaches for interpreting deep learning models in clinical applications.

## 4.1 Limitations

Despite the advantages of our approach, several limitations should be acknowledged. First, the optimal number of principal components for correction was determined through visual assessment of localization, introducing a degree of subjectivity. Future work could develop more quantitative criteria for selecting the optimal correction level.

Second, while our approach creates targets with known anatomical localization, it does not fully capture the distributed nature of many neurological conditions. Real disease patterns often involve networks of regions with complex interactions, which are not directly represented by our single-region targets. However, our approach could be extended to create multi-region targets by combining corrected IDPs.

Third, the transformation from raw to corrected IDPs changes the semantic meaning of the prediction targets. The corrected version represents a more abstract measure of e.g. local hippocampal morphology independent of global brain characteristics, and thus cleaned of e.g. age and sex confounds and therefore should more directly relate to clinical conditions like Alzheimer's disease. Still, this transformation should be considered with some caution when interpreting the clinical relevance of model predictions and explanations.

Finally, our approach currently focuses on structural MRI data and may not generalize directly to functional imaging modalities or multimodal integration, which present additional challenges for XAI validation.

## 4.2 Future Directions

Several promising directions emerge from this work. The framework could be extended to more complex scenarios by creating composite targets representing distributed patterns, similar to disease signatures. This would enable validation of XAI methods for detecting patterns that span multiple brain regions with varying strengths.

The approach could also be applied to longitudinal data, creating targets that represent region-specific changes over time. This would address the critical need for validating explanations of predictive models for disease progression.

Integration with causal modeling approaches could further strengthen the validation framework by distinguishing between direct causal relationships and indirect correlations in explanations. This would be particularly valuable for clinical applications where understanding causal mechanisms is essential.

Automated optimization of the correction procedure, potentially through quantitative metrics of localization quality, would enhance reproducibility and reduce the subjective elements of the current approach.

Finally, extending the framework to other imaging modalities, such as functional MRI or diffusion tensor imaging, would broaden its applicability to diverse neuroscientific questions.

## 5. Conclusion

We have introduced a novel approach for creating anatomically localized prediction targets from imaging-derived phenotypes, enabling objective validation of explainable AI methods in neuroimaging. By systematically removing global brain effects through principal component analysis, we generate targets with verifiable spatial localization that remain learnable by deep neural networks. The effectiveness of this approach has been demonstrated across diverse brain measures, including subcortical intensities, regional volumes, cortical thickness, and surface areas.

This framework addresses a critical gap in the field by providing ground truth for model explanations, facilitating systematic evaluation and comparison of different XAI methods. The implications extend beyond methodological validation to clinical applications, where reliable interpretation of model decisions is essential for responsible deployment of AI in healthcare.

As deep learning continues to advance in neuroimaging applications, frameworks for ensuring the interpretability and trustworthiness of these models become increasingly important. Our approach represents a significant step toward bridging the interpretability gap, potentially accelerating the translation of AI advances into clinical practice while maintaining the scientific rigor necessary for neuroimaging research.

**Appendix: Detailed Methodology**

**A.1 Data Processing and Correlation Analysis**

The UK Biobank dataset provides T1-weighted structural MRI scans for approximately 46,000 participants, acquired using a standard Siemens Skyra 3T scanner with the following parameters: 1×1×1mm resolution (Alfaro-Almagro et al., 2018). Images were preprocessed by the UK Biobank imaging team with their standard pipeline, including gradient distortion correction, field of view reduction, and registration to standard space.

For our correlation analysis to visualize IDP localization, we processed the images as follows: T1-weighted images were linearly registered to MNI space (resolution: 182×218×182), downsampled to half resolution (91×109×91) using local mean pooling to reduce memory requirements for the subsequent statistical analysis, smoothed with a Gaussian kernel (FWHM=2 at half resolution, equivalent to 4mm at original resolution), and masked with a brain mask derived from 10,000 images. The voxel-wise correlation analysis was performed using a permutation-based Ordinary Least Squares approach implemented in nilearn, using 5,000 images, 200 permutations, and a static bias as the only confounding variable. The resulting negative log p-values were signed according to t-values and visualized (where neg_log_p = 1.3 corresponds to p = 0.05). In the visualization, we outlined the target anatomical region from the Destrieux atlas (which combines cortical parcellation with subcortical segmentation), dilated by 2mm to account for the importance of boundary voxels, particularly for volumetric measures.

**A.2 IDP Selection and Correction**

The full list of 10 selected IDPs with their UK Biobank field IDs is provided in Table A1. For each target IDP, we constructed a correction set excluding all related measurements as described in the Methods section.

Table A1: Selected IDPs with their UK Biobank field IDs and descriptions.

| IDP Name | UK Biobank Field ID | Description |
| --- | --- | --- |
| Mean intensity of Pallidum (right hemisphere) | 26576.2.0 | Mean intensity of Pallidum in the right hemisphere generated by subcortical volumetric segmentation (aseg) |
| Mean intensity of Putamen (left hemisphere) | 26544.2.0 | Mean intensity of Putamen in the left hemisphere generated by subcortical volumetric segmentation (aseg) |
| Volume of Hippocampus (left hemisphere) | 26562.2.0 | Volume of Hippocampus in the left hemisphere generated by subcortical volumetric segmentation (aseg) |
| Volume of Caudate (left hemisphere) | 26559.2.0 | Volume of Caudate in the left hemisphere generated by subcortical volumetric segmentation (aseg) |
| Mean thickness of G-postcentral (right hemisphere) | 27652.2.0 | Mean thickness of G-postcentral in the right hemisphere generated by parcellation of the white surface using Destrieux (a2009s) parcellation |
| Mean thickness of G-insular-short (left hemisphere) | 27420.2.0 | Mean thickness of G-insular-short in the left hemisphere generated by parcellation of the white surface using Destrieux (a2009s) parcellation |
| Area of G-rectus (left hemisphere) | 27359.2.0 | Area of G-rectus in the left hemisphere generated by parcellation of the white surface using Destrieux (a2009s) parcellation |
| Area of G-orbital (left hemisphere) | 27352.2.0 | Area of G-orbital in the left hemisphere generated by parcellation of the white surface using Destrieux (a2009s) parcellation |
| Volume of Brain-Stem (whole brain) | 26526.2.0 | Volume of Brain-Stem in the whole brain generated by subcortical volumetric segmentation (aseg) |
| Volume of Lateral-Ventricle (left hemisphere) | 26554.2.0 | Volume of Lateral-Ventricle in the left hemisphere generated by subcortical volumetric segmentation (aseg) |

The principal component analysis was performed using scikit-learn's PCA implementation with standardized inputs. For each target IDP, we computed corrections using increasing numbers of principal components (0, 5, 10, 15, etc.) and assessed localization through correlation maps. For subcortical intensity targets used in initial method development, we used a finer increment (1 PC at a time) to carefully evaluate the progression of localization.

### A.3 Model Architecture and Training

For image preprocessing in our deep learning pipeline, we used the minimally preprocessed T1-weighted MRI scans provided by the UK Biobank, which were skull-stripped with the UK Biobank-provided brain mask and linearly registered to MNI152 space using the UK Biobank-provided transformation matrices. Images were normalized using constant values derived from the training set (mean = 232.55, std = 414.41).

We used a 3D ResNet architecture with 18 layers, implemented in PyTorch. The network accepts the preprocessed brain images as input and produces a single continuous output prediction. The ResNet blocks follow the standard implementation with 3D convolutions, batch normalization, and ReLU activations.

Training was conducted using the PyTorch Lightning framework on Nvidia A100 GPUs with 80GB memory. We used the Adam optimizer with a one-cycle learning rate policy, maximum learning rate of $10^{-4}$ (lower than the $10^{-2}$ used in our brain age work to ensure learnability across all IDP targets), and mean squared error loss function. Training proceeded for 150,000 gradient updates with a batch size of 8, which typically required approximately 2 days per model.

The corrected IDP targets were standardized to zero mean with unit variance for training stability, with the original standard deviations shown in Table A3. The training, validation, and test splits were created randomly at the subject level, ensuring that no subject appeared in multiple splits.

## A.4 Region Masking Validation

For the region masking experiments, we used the Destrieux atlas to define anatomical masks for each target region. These masks were dilated by 2mm to include boundary voxels that might be particularly informative for volume-based measures. The masks were applied to the input images by setting all voxels within the masked region to zero, while leaving the rest of the image unchanged.

Models were then trained on these masked images following the same protocol as the main analysis, and their performance was evaluated on similarly masked test images. This approach isolates the effect of the target region on prediction performance by comparing models trained on full brain images versus those with the target region removed.

## A.5 Full Results Tables

Table A2 provides the complete results of our analysis, including the IDP type, optimal number of principal components for each target IDP, the model performance on both the original and corrected targets, and the performance drop observed in the region masking validation.

Table A2: Detailed results of IDP correction, model performance, and region masking validation.

| IDP Target | IDP Type | Optimal n_PC | $R^2$ (uncorrected)[1] | $R^2$ (corrected) | $R^2$ (region masked) | Performance drop | % of significantly correlated voxels in target mask (uncorrected) | % of significantly correlated voxels in target mask (corrected) | % gained |
|---|---|---|---|---|---|---|---|---|---|
| Mean intensity of Pallidum (right) | Subcortical intensity | 75 | 0.62 | 0.74 | 0.00 | 0.74 | 0.15 | 1.00 | 0.85 |
| Mean intensity of Putamen (left) | Subcortical intensity | 20 | 0.77 | 0.74 | 0.00 | 0.74 | 0.16 | 0.97 | 0.81 |
| Volume of Hippocampus (left) | Subcortical Volume | 45 | 0.83 | 0.72 | 0.04 | 0.68 | 0.11 | 0.90 | 0.79 |
| Volume of Caudate (left) | Subcortical Volume | 25 | 0.89 | 0.88 | 0.03 | 0.85 | 0.14 | 0.99 | 0.85 |
| Mean thickness of G-postcentral (right) | Cortical thickness | 100 | 0.56 | 0.51 | 0.00 | 0.51 | 0.08 | 0.60 | 0.52 |
| Mean thickness of G-insular-short (left) | Cortical thickness | 325 | 0.52 | 0.27 | 0.00 | 0.27 | 0..07 | 0.52 | 0.45 |
| Area of G-rectus (left) | Cortical area | 425 | 0.66 | 0.41 | 0.00 | 0.41 | 0.06 | 0.97 | 0.91 |
| Area of | Cortical | 145 | 0.81 | 0.46 | 0.00 | 0.46 | 0.13 | 0.96 | 0.83 |

---

[1] Performance for uncorrected IDPs was computed on the validation set.

| | | | | | | | | |
|---|---|---|---|---|---|---|---|---|
| G-orbital area (left) | | | | | | | | |
| Volume of Lateral-Ventricle (left) | Volume | 55 | 0.91 | 0.71 | 0.00 | 0.71 | 0.42 | 0.97 | 0.55 |
| Volume of Brain-Stem (whole) | Volume | 15 | 0.99 | 0.77 | 0.31 | 0.46 | 0.10 | 0.45 | 0.35 |

## A.6 Explanation Postprocessing

For the explanation evaluation, we processed the attribution maps using a standardized pipeline to ensure fair comparisons across methods and subjects:

1) Taking the absolute values of attributions since we are interested in magnitude rather than direction for regression tasks
2) Applying spatial smoothing with FWHM = 4mm (equivalent to FWHM = 2 at half resolution) to reduce the penalty for activations just outside the ground truth mask
3) Scaling to the 99th percentile to make explanations comparable across subjects and methods
4) Applying a cutoff at the 99th percentile to reduce visual noise

The target region masks used for explanation evaluation were dilated by 2mm to account for the importance of boundary information. For quantitative evaluation metrics, we used relevance mass accuracy (Arras et al., 2022) to assess how well the explanation aligns with the known anatomical ground truth.

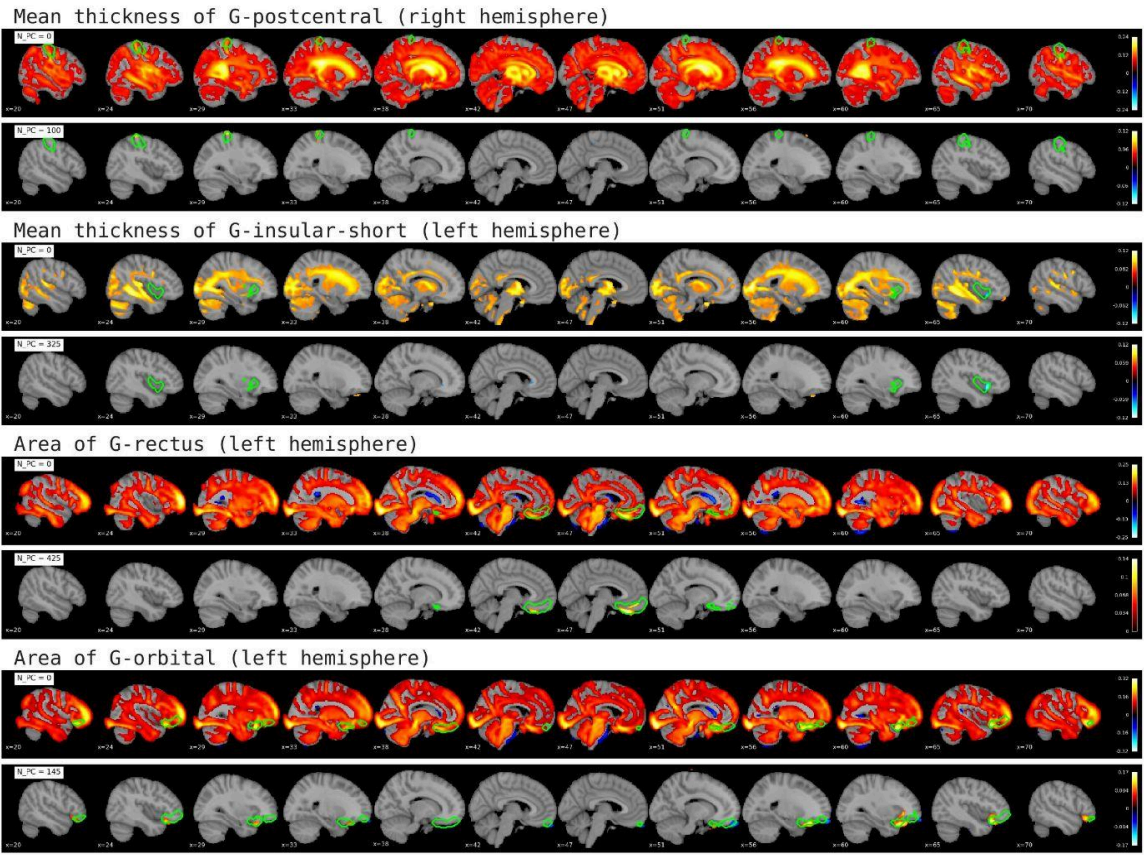

Figure A1a: *Anatomical localization of cortical IDP targets before and after correction. Each row shows a correlation map between individual voxels and a single (corrected) IDP, with the associated anatomical region (dilated by 2 mm) outlined in green on sagittal brain slices. Brighter colors represent stronger effects. Correlation maps are masked by FWE-corrected significance (α = 0.05), as computing high-resolution p-maps was computationally infeasible.*
*The consistent anatomical localization across targets highlights the robustness of our correction procedure for varying cortical thickness and area measurements.*

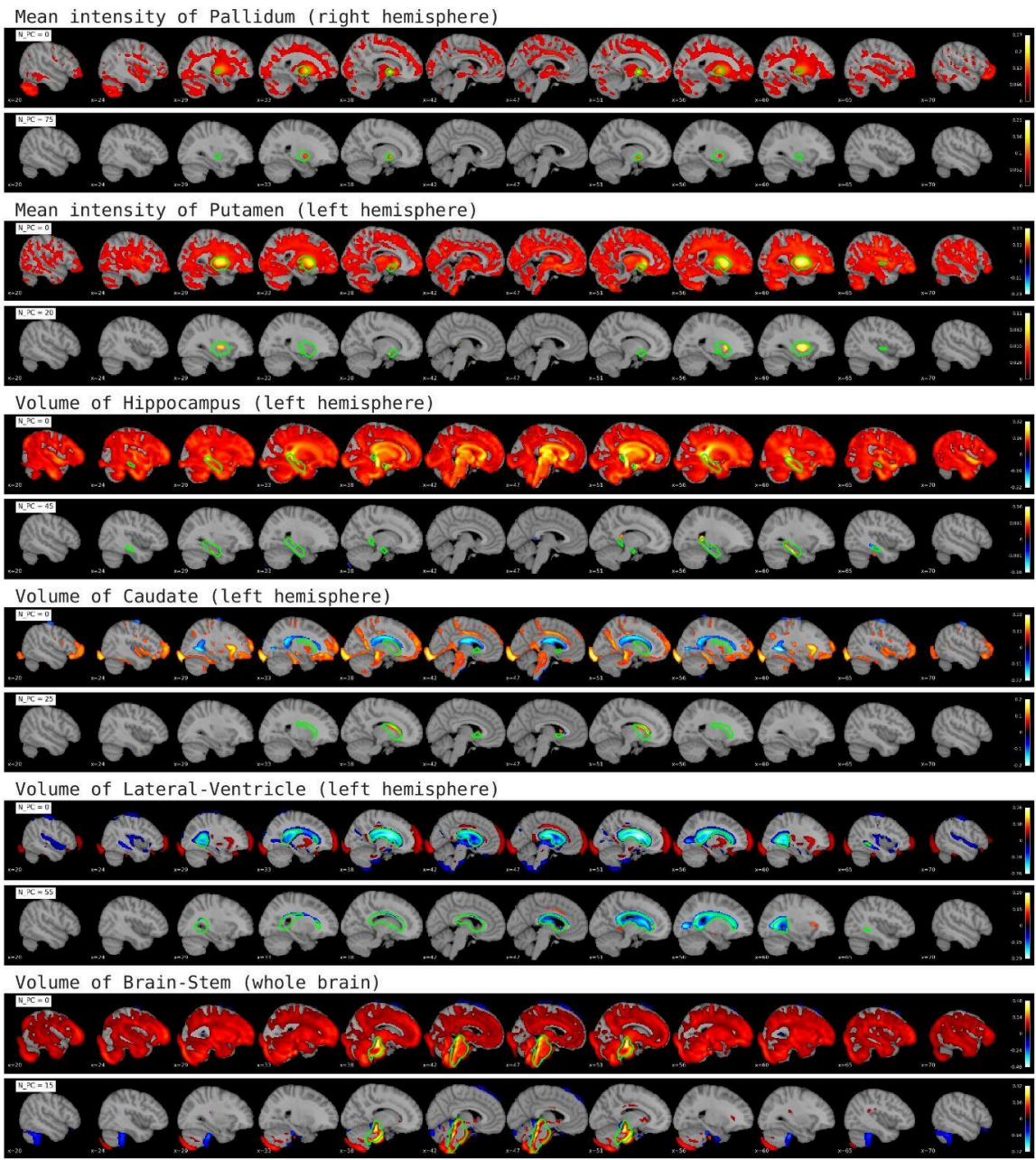

*Figure A2b: Localization of corrected subcortical IDP targets analogous to Figure A1a.*

## Acknowledgments

This research was conducted using the UK Biobank Resource under Application Number 33073. We thank the participants and researchers who made this dataset possible. Computational resources were provided by the High Performance Computing infrastructure at Charité - Universitätsmedizin Berlin. The manuscript was edited for grammar and style using LLMs.

# References

Alfaro-Almagro, F., Jenkinson, M., Bangerter, N. K., Andersson, J. L. R., Griffanti, L., Douaud, G., Sotiropoulos, S. N., Jbabdi, S., Hernandez-Fernandez, M., & Vallee, E. (2018). Image processing and Quality Control for the first 10,000 brain imaging datasets from UK Biobank. *Neuroimage*, *166*, 400–424.

Arras, L., Osman, A., & Samek, W. (2022). CLEVR-XAI: A benchmark dataset for the ground truth evaluation of neural network explanations. *An International Journal on Information Fusion*, *81*, 14–40.

Arun, N., Gaw, N., Singh, P., Chang, K., Aggarwal, M., Chen, B., Hoebel, K., Gupta, S., Patel, J., Gidwani, M., Adebayo, J., Li, M. D., & Kalpathy-Cramer, J. (2021). Assessing the trustworthiness of saliency maps for localizing abnormalities in medical imaging. *Radiology. Artificial Intelligence*, *3*(6), e200267.

Budding, C., Eitel, F., Ritter, K., & Haufe, S. (2021). Evaluating saliency methods on artificial data with different background types. In *arXiv [eess.IV]*. arXiv. http://arxiv.org/abs/2112.04882

Hedges, E. P., Dimitrov, M., Zahid, U., Brito Vega, B., Si, S., Dickson, H., McGuire, P., Williams, S., Barker, G. J., & Kempton, M. J. (2022). Reliability of structural MRI measurements: The effects of scan session, head tilt, inter-scan interval, acquisition sequence, FreeSurfer version and processing stream. *NeuroImage*, *246*(118751), 118751.

Hofmann, S. M., Beyer, F., Lapuschkin, S., Goltermann, O., Loeffler, M., Müller, K.-R., Villringer, A., Samek, W., & Witte, A. V. (2022). Towards the interpretability of deep learning models for multi-modal neuroimaging: Finding structural changes of the ageing brain. *NeuroImage*, *261*(119504), 119504.

Kelly, C. J., Karthikesalingam, A., Suleyman, M., Corrado, G., & King, D. (2019). Key challenges for delivering clinical impact with artificial intelligence. *BMC Medicine*, *17*(1), 195.

Litjens, G., Kooi, T., Bejnordi, B. E., Setio, A. A. A., Ciompi, F., Ghafoorian, M., van der Laak, J. A. W. M., van Ginneken, B., & Sánchez, C. I. (2017). A survey on deep learning in medical image analysis. *Medical Image Analysis*, *42*, 60–88.

Mechelli, A., Friston, K. J., Frackowiak, R. S., & Price, C. J. (2005). Structural covariance in the human cortex. *The Journal of Neuroscience: The Official Journal of the Society for Neuroscience*, *25*(36), 8303–8310.

Molnar, C., Casalicchio, G., & Bischl, B. (2020). Interpretable machine learning – A brief history, state-of-the-art and challenges. In *Communications in Computer and Information Science* (pp. 417–431). Springer International Publishing.

Montavon, G., Samek, W., & Müller, K.-R. (2018). Methods for interpreting and understanding deep neural networks. *Digital Signal Processing*, *73*, 1–15.

Oliveira, M., Wilming, R., Clark, B., Budding, C., Eitel, F., Ritter, K., & Haufe, S. (2024). Benchmarking the influence of pre-training on explanation performance in MR image classification. *Frontiers in Artificial Intelligence*, *7*, 1330919.

Ras, G., Xie, N., Van Gerven, M., & Doran, D. (2022). Explainable deep learning: A field guide for the uninitiated. *The Journal of Artificial Intelligence Research*, *73*, 329–397.

Schulz, M.-A., Siegel, N. T., & Ritter, K. (2024). Beyond accuracy: Refining brain-age models for enhanced disease detection. In *bioRxiv* (p. 2024.03. 28.587212). https://doi.org/10.1101/2024.03.28.587212

Shen, D., Wu, G., & Suk, H.-I. (2017). Deep learning in medical image analysis. *Annual Review of Biomedical Engineering*, *19*, 221–248.

Siegel, N. T., Kainmueller, D., Deniz, F., Ritter, K., & Schulz, M.-A. (2025). Do transformers and CNNs learn different concepts of brain age? *Human Brain Mapping*, *46*(8). https://doi.org/10.1002/hbm.70243

**SM-A: Validation Framework: Methodology and Ground Truth Establishment**

**(A1) Framework Overview & Rationale**

To overcome the challenge of evaluating XAI methods without access to the model's internal "true" reasoning, we developed a validation framework specifically designed for the complexities of 3D neuroimaging. This framework allows us to systematically assess XAI method reliability against verifiable ground truth across tasks of increasing complexity, using large-scale, realistic brain MRI data. The core principle is to establish ground truth by manipulating the prediction target, rather than altering the input images themselves. This approach is critical because it ensures the validation is performed on unmodified, in-distribution data. Modifying input images, for instance by inserting synthetic lesions, introduces a risk that any observed effects are due to artifacts from the image modification itself, rather than the method's ability to identify genuine features. By keeping the images pristine, we can be confident that our findings reflect a method's performance on realistic data and are more likely to translate to real-world applications.

The design of the four stages follows a deliberate gradient of increasing complexity, which is essential for bridging the gap between experimental control and real-world applicability. Stage 1 (cIDPs) provides maximal control with unambiguous ground truth, but these single-region targets represent a simplification of complex neurobiology. Therefore, we progressively introduce more realistic scenarios: Stages 2 and 3 test for distributed patterns, while Stage 4 assesses plausibility in a highly complex setting where spatial ground truth is unavailable. The central strength of this design is that if an XAI method's performance is consistent across this entire gradient - from the most controlled tasks to the most realistic - we can more confidently infer that its explanations will be reliable in real-world applications where the ground truth is fundamentally unknown. The framework comprises four stages, each detailed in the following sections (SM-A3 to SM-A6).

**(A2) Dataset Details**

All neuroimaging data were obtained from the UK Biobank (UKBB) resource under Application Number 33073. The UKBB has ethical approval from the North West Multi-centre Research Ethics Committee. After quality control, our dataset comprised scans from 46,381 participants. For Stages 1-4, we used T1-weighted structural MRI scans acquired on a standard Siemens Skyra 3T scanner (1x1x1mm resolution). For Stage 3 (Lesion Prediction), we used T2-weighted scans.

For our deep learning pipeline, we used the minimally preprocessed T1-weighted MRI scans provided by the UK Biobank. These were skull-stripped and linearly registered to MNI152 space using transformations provided by the UKBB. Images were then normalized using constant values derived from the training set.

**(A3) Stage 1 - Corrected IDP Generation and Validation**

Stage 1 establishes ground truth using anatomically localized targets derived from Imaging-Derived Phenotypes (IDPs). The complete methodology for generating and validating these targets is detailed in our `accompanying cIDP methods manuscript`; the following provides a summary.

Raw IDPs (e.g., regional volume) often correlate with global brain features, making them unsuitable for localized ground truth. We developed a novel correction procedure to address this.

*Methodology*: We selected 10 diverse IDPs representing subcortical intensities, regional volumes, cortical thickness, and surface areas. To remove confounding global effects, we applied a Principal Component Analysis (PCA)-based correction. For each target IDP (e.g., Hippocampus volume), we constructed a large set of related cortical or subcortical IDPs, excluding the target itself. We performed PCA on this set to capture major sources of shared variance. The raw target IDP was then regressed against these principal components, and the residuals were retained as the corrected IDP (cIDP). This process yields a prediction target driven by local anatomy rather than global factors. The optimal number of PCs to remove was determined by visually assessing voxel-wise correlation maps to ensure the signal was precisely localized to the anatomical region of interest.

*Validation*: We validated the localization of our cIDPs using three approaches. First, we visually confirmed that voxel-wise correlations between the cIDP and the brain image were tightly focused on the target region. Second, using a mass-univariate approach, we assessed the proportion of significant voxel-wise correlations (alpha = 0.05) falling within the target mask - confirming our visual impression. Third, we conducted a critical masking experiment: we trained models to predict the cIDP from images where the target anatomical region was masked out (i.e., set to zero). For all cIDPs, prediction performance ($R^2$) dropped to near-zero, confirming that the model relied specifically on the target region's anatomy. For example, a model predicting the corrected mean intensity of the pallidum achieved an $R^2$ of 0.74 with full brain images, which dropped to 0.00 when the pallidum was masked. Minor exceptions were noted for the brain stem and ventricles, where some residual predictability remained due to their extensive connections and large, defined shape.

This process ensures that the cIDPs serve as valid, spatially precise ground truth targets for evaluating XAI explanations. For a comprehensive description of the methodology, the full list of IDPs, and detailed validation results, please refer to the accompanying cIDP methods manuscript.

*Causal Structure:* Our correction approach alters the causal structure between the prediction target, the target region, and non-target brain regions. In the uncorrected setting (i.e. using standard IDP as prediction targets), the target region has a direct causal link to the target variable. This means that the model can, in principle, rely on the target region alone for its predictions, without needing context information (so-called suppressor variables; Wilming et al. (2022)) from other brain areas. However, real-world applications typically require interpretation relative to such suppressor variables. For instance, brain region volumes must often be contextualized against global atrophy due to aging. While in theory, the absence of suppressor variables enables the model to focus solely on the target region, in practice, the high inter-regional correlations in brain structure may lead the model to rely on proxy features instead. For example, when predicting the volume of a small, furrowed subcortical region, a typical CNN might learn to use more visually salient structures like the ventricles, especially when training data is limited. This possibility to rely on proxy features undermines the suitability of raw IDPs as prediction targets for ground-truth XAI validation, since it creates an ambiguity on whether attributions outside the target region stem from XAI failures or models relying on proxy features for their predictions.

Using cIDPs instead of raw IDPs prevents models from relying on proxy features and offers a more realistic testbed for XAI validation, as predicting cIDPs requires interpretation relative to suppressor variables. To accurately predict a cIDP, the model must estimate properties of a target region in relation to the global brain effects removed during correction. This creates a plausible source for relevance attributions outside the target-regions mask, potentially reflecting suppressor variables. However, we argue that suppressor variables are a rather technical insight of model behavior and in biomedical contexts, XAI methods should prioritize highlighting regions that are directly and meaningfully related to the prediction target.

**(A4) Stage 2 - Controlled Distributed Patterns ("Artificial Diseases")**

To evaluate whether explainable AI (XAI) methods are sensitive to distributed predictive information in the brain, we introduce *artificial diseases* - synthetic binary classification targets defined by combinations of corrected imaging-derived phenotypes (cIDPs) from multiple regions. Importantly, this design models a disease classification scenario, where diagnostic decisions often rely on local abnormalities across multiple brain regions. For example, Alzheimer's disease may involve atrophy in both the Hippocampus and Frontal Lobes (Kang et al., 2024), requiring interpretation against a background of global variation due to factors like aging.

|  | cIDP 1 | cIDP 2 |
|---|---|---|
| Artificial Disease 1 | Area of the Gyrus Rectus | Volume of the Caudate |
| Artificial Disease 2 | Mean Thickness of the Postcentral Gyrus | Volume of the Hippocampus |

Table SM-A.T1: Specific cIDPs used for generating artificial diseases (for details on each cIDP see SM-A3).

We created binary labels for two artificial diseases by combining a cortical cIDP with a subcortical cIDP for each disease (Table SM-A.T1), reflecting the realistic scenario in which clinical conditions affect both cortical and subcortical brain regions. We assigned a patient label (1) to all subjects where cIDP 1 was larger than the 60th percentile and cIDP 2 was smaller than or equal to the 40th percentile, thereby capturing co-occurring deviations in both modalities ($n_{patients\_artificial\_disease\_1}$ = 7,407, $n_{patients\_artificial\_disease\_2}$ = 7,258). To sharpen the decision boundary and reduce ambiguity, we excluded subjects with cIDP values between the 40th and 60th percentiles for either cIDP. This filtering step improved class separation and yielded a final dataset representing approximately 64% of the original cohort ($n_{total\_artificial\_disease\_1}$ = 29,246, $n_{total\_artificial\_disease\_2}$ = 29,268). Remaining subjects not labeled as patients were considered controls ($n_{controls\_artificial\_disease\_1}$ = 21,839, $n_{controls\_artificial\_disease\_2}$ = 22,010). Although this led to an imbalanced dataset with an approximate 1:3 ratio of patients to controls, our 3D ResNet-50 achieved accuracies greater than 0.75 and non-zero precision for both artificial diseases, indicating successful learning of the underlying signal rather than a trivial bias toward the majority class ($accuracy_{artificial\_disease\_1}$ = 0.83, $accuracy_{artificial\_disease\_2}$ = 0.80; detailed performance metrics in ST-1). The model training procedure, including the specific splits used for training, validation, and testing, is described in SM-B.

Finally, to evaluate explanations for models trained on artificial diseases, we combined the ground truth masks corresponding to each cIDP involved in a given artificial disease into a single composite ground truth mask. Results from this evaluation are presented in Figures 1.c and 5.5, as well as in Supplementary Material SM-C.

**(A5) Stage 3 - Lesion Prediction Task Setup**

To evaluate XAI methods on clinically relevant and heterogeneous distributed patterns, we investigate models trained to predict the volume of individual white matter hyperintensity (WMH) lesion load - a clinical neuroimaging marker associated e.g. with cognitive decline, stroke, multiple sclerosis, dementia, and death (Debette & Markus, 2010; Di Stadio et al., 2018). Unlike the synthetic targets in Stage 2, WMH lesions reflect real pathological changes with spatially variable, multifocal distributions. Ground truth lesion segmentations, automatically derived using the BIANCA tool (Griffanti et al., 2016) and provided by the UK Biobank (UKBB), enable systematic evaluation of explanation methods against clinically meaningful and anatomically distributed effects.

For model training and evaluation, we used 44,990 T2-weighted MRI scans from the UKBB, with the "Total volume of periventricular white matter hyperintensities derived from T1 and T2-FLAIR images" (UKBB field 24485.2.0) as the prediction target (details about model architecture and training protocol in SM-B). Our 3D ResNet-50 demonstrated strong predictive performance ($R^2$ = 0.93; further metrics in Supplementary Table ST-1). For evaluating XAI methods, we registered the ground truth WMH segmentations to MNI space, aligning with the linear registration of input images. To generate the group-level visualization in Figure 5.6, we selected subjects with lesion loads above the 99th percentile, computed their mean explanation maps and corresponding mean lesion masks, binarized the latter using a 0.5 threshold, and scaled the mean explanation maps by a factor of 4 for visual clarity.

## (A6) Stage 4 - Brain Age Plausibility Analysis

In Stage 4, we evaluate explanation methods in a setting where biological complexity is high but spatial ground truth is inherently unavailable: brain age prediction. Predicting chronological age from structural brain MRI is a widely used biomarker in neuroimaging, where deep learning models are known to excel (Leonardsen et al., 2022; Peng et al., 2021; Siegel et al., 2025). Yet the absence of direct spatial targets poses a challenge for validating explanation quality. To address this, we introduce a principled, literature-driven plausibility check: we quantify how well the spatial relevance patterns identified by different XAI methods align with neuroanatomical signatures of aging reported in large-scale meta-analysis. This allows us to assess whether models rely on biologically meaningful features rather than arbitrary or spurious correlates - an essential step toward building trust in explanations for complex, distributed biomarkers, where predictive signals are often highly intercorrelated across brain regions (Bethlehem et al., 2022).

To perform this comparison, we drew on established anatomical findings from meta-analysis, specifically Walhovd et al. (2011), which reported consistent age-related structural differences across multiple cortical and subcortical regions. We used this literature-based set of 17 regions (Cerebral Cortex, Cerebral White Matter, Lateral Ventricle, Inferior Lateral Ventricle, Cerebellum White Matter, Cerebellum Cortex, Thalamus, Caudate, Putamen, Pallidum, Hippocampus, Amygdala, Accumbens Area, Third Ventricle, Fourth Ventricle, Brain Stem, CSF) as a reference for evaluating the plausibility of model explanations.

We trained a 3D ResNet-50 for brain age prediction using T1-weighted MRI scans from the UK Biobank (UKBB), following standard protocols. The model was trained on healthy control subjects (n = 27,513) and evaluated on a held-out set of healthy individuals (n = 1,172). For specific training details, see SM-B. Explanations were generated for each held-out healthy subject, each XAI method, and three different model initializations. Spatial relevance was then assessed at the level of predefined anatomical regions. We considered 191 regions from the Destrieux atlas (Destrieux et al., 2010) and Freesurfer ASEG subsegmentations (Fischl et al., 2002), treating left and right hemispheres separately. To account for anatomical variability across individuals, we used deformation fields provided by the UKBB to inverse-warp each region from nonlinear MNI space into each participant's linear MNI space. This yielded accurate, participant-specific binary region masks that improved spatial alignment between region mask and anatomical structure.

Each brain region was assigned a relevance score based on the 99th percentile of explanation values within the participant-specific mask. This percentile-based scoring approach provided a balanced measure of regional relevance: it avoided the underestimation of large regions that may have contained localized but meaningful signal, while also reducing the risk of small regions being overemphasized due to spurious high explanation values falling within their limited spatial extent. For regions with bilateral counterparts, we retained only the higher-scoring hemisphere, yielding a final ranking of 101 unique regions per participant.

We compared each participant's top 17 regions to the 17 regions identified in the meta-analysis by Walhovd et al. (2011), using the proportion of overlap as a simple yet informative metric of biological plausibility. This comparison was performed separately for each XAI method and model initialization. For each participant, we averaged the overlap scores across the independent model initializations. Final results were reported as the population mean and standard deviation across all participants. This procedure yielded a robust, quantitative measure of whether explanation methods consistently reflected known patterns of brain aging, offering critical insight into the biological validity of model explanations in the absence of spatial ground truth. We reported the final overlap scores in ST-2 and the population standard deviation in ST-3.

## SM-B: Deep Learning Model and XAI Implementation Details

### (B1) Model Architecture

We perform all experiments using the well-established ResNet-50 architecture (He et al., 2016), applying a 3D adaptation for brain imaging tasks (Hara et al., 2018). ResNet is among the most widely used deep learning model architectures in neuroimaging domains (Abrol et al., 2020; Chatterjee et al., 2022; Jonsson et al., 2019; Kolbeinsson et al., 2020; Younis et al., 2024), making it a representative choice for evaluating XAI methods in settings that reflect common practice. We use a standard PyTorch implementation[1].

### (B2) Model Training

The training protocol was consistent across Stages 1 to 3 (cIDPs, artificial diseases, lesions) and largely followed our earlier work on brain age prediction from 3D MRI scans (Schulz et al., 2024; Siegel et al., 2025). The main adjustment was a reduction of the maximum learning rate from 0.01 to 0.0001 to ensure stable convergence across all cIDP prediction targets. For the brain age plausibility analysis (Stage 4), the original learning rate (0.01) was retained. MRI images were standardized using neuroimaging-specific constants, and scalar prediction targets were z-scored based on the mean and standard deviation computed from the training set. All regression models - including those predicting cIDPs, lesion volume, and brain age - were trained using mean squared error loss. For binary classification tasks (artificial diseases), we applied binary cross-entropy loss and modified the regression head by appending a sigmoid activation to the final output logit. Training was conducted using PyTorch 1.12 and PyTorch Lightning 1.8 on two NVIDIA A100 GPUs, with the Adam optimizer (Kingma & Ba, 2014) and a one-cycle learning rate policy (Smith & Topin, 2019). Each model was trained for 150,000 gradient update steps with a batch size of 8.

Data splits for training, validation, and testing are summarized in Supplementary Material Table SM-B.T1. The validation set was used to confirm and tune hyperparameters - most importantly, to select an appropriate learning rate - while the test set was reserved for reporting predictive performance and for evaluating model explanations. The splits were generated by the following procedure: for cIDP prediction tasks, we defined a base cohort of 45,760 subjects with available T1-weighted images and complete subcortical IDPs, randomly split into 80% training, 10% validation, and 10% test sets. For cortical targets, artificial disease classification, and WMH lesion prediction, a small number of subjects were excluded from the base cohort due to missing relevant imaging data (e.g., unavailable cortical IDPs or missing T2-weighted scans). In the artificial disease setting, we additionally removed subjects near the decision boundary to sharpen class separation. These exclusions were performed within the existing split structure to preserve cohort comparability across tasks. For brain age prediction, we followed the healthy-train/healthy-test design established in prior work (Schulz et al., 2024; Siegel et al., 2025), using only participants without neurological or psychiatric diagnoses. The T2-based brain age splits were derived by filtering the T1-based brain age cohort to retain only subjects with available T2-weighted scans.

---

[1] https://github.com/kenshohara/3D-ResNets-PyTorch

| Task | N Training | N Validation | N Test | N Total |
|---|---|---|---|---|
| Subcortical Targets | 36608 | 4576 | 4576 | 45760 |
| Cortical Targets | 36575 | 4574 | 4571 | 45720 |
| Artificial Disease 1 | 23385 | 2938 | 2923 | 29246 |
| Artificial Disease 2 | 23438 | 2909 | 2921 | 29268 |
| Lesions | 35533 | 4442 | 4436 | 44411 |
| Brain AGE T1 | 27513 | 17685 | 1172 | 46370 |
| Brain AGE T2 | 26686 | 17167 | 1138 | 44991 |

Table SM-B.T1: Number of subjects used for model training, evaluation and testing for the different

Predictive performance metrics for all tasks are reported in Table ST-1. For regression tasks (including cIDP prediction, lesion load, and brain age), we report mean absolute error (MAE), standard deviation of MAE across test subjects (MAE STD), and coefficient of determination ($R^2$). For binary classification tasks (artificial diseases), we report accuracy, precision, and recall. These metrics confirm successful training across tasks and provide a quantitative foundation for interpreting model behavior and explanation quality.

**(B3) XAI Method Implementation**

We benchmarked a comprehensive set of XAI methods[2], including Layer-wise Relevance Propagation (LRP), GradCAM, SmoothGrad, DeepLift, Guided Backpropagation, Excitation Backprop, and Input × Gradient. These methods represent the most widely used XAI techniques in neuroimaging analysis with deep learning (see SM-G), spanning the major conceptual classes of feature attribution: gradient-based, relevance-based, reference-based, and CAM-based methods. Implementation details and parameter choices for each method are described below.

**LRP Rulesets**
We implemented several LRP variants using the *Zennit* library (v0.5.1) (Anders et al., 2021), which was developed by researchers closely associated with the original LRP framework (Bach et al., 2015). We selected composite rule sets - LRP_EpsilonAlpha2Beta1, LRP_EpsilonAlpha2Beta1Flat, LRP_EpsilonPlus, and LRP_EpsilonPlusFlat (see Table SM-B.T2 for an overview) - and used Zennit's default parameter settings (e.g., epsilon values) for each rule. The use of composite LRP rules has emerged as best practice, as they have been shown to more accurately reflect the model's reasoning and improve object localization based on human evaluations (Kohlbrenner et al., 2020). We applied all LRP variants using the default ResNet canonization provided by Zennit.

---

[2] We refer to feature/relevance attribution methods as XAI methods throughout our study.

| LRP Variant | First Layer | Middle layers | Last, fully-connected Layer |
|---|---|---|---|
| LRP_EpsilonAlpha2Beta1 | LRP_Alpha2Beta1 | LRP_Alpha2Beta1 | LRP_Epsilon |
| LRP_EpsilonAlpha2Beta1 Flat | LRP_Flat | LRP_Alpha2Beta1 | LRP_Epsilon |
| LRP_EpsilonPlus | LRP_ZPlus | LRP_ZPlus | LRP_Epsilon |
| LRP_EpsilonPlusFlat | LRP_Flat | LRP_ZPlus | LRP_Epsilon |

Table SM-B.T2: Different propagation rules used in each LRP variant.

Further, we benchmarked Input × Gradient (Shrikumar et al., 2017) and Excitation Backprop (Zhang et al., 2018), see below), both of which can be interpreted as specific LRP variants. Input × Gradient corresponds to applying the LRP_0 rule across all network layers (cf. Shrikumar et al., 2017), while Excitation Backprop is equivalent to applying LRP_ZPlus throughout the network, i.e., propagating only positive contributions.

**GradCAM**
GradCAM (Selvaraju et al., 2017) was implemented using the widely adopted *pytorch-grad-cam* library (v1.5.4). As a default, we extracted activations from the final convolutional layer following standard practice. To examine the effect of spatial resolution, we also generated GradCAM maps using activations from the last convolutional layer of the third layer group, yielding higher-resolution heatmaps.

For completeness, additional GradCAM variants were computed using activations from the last convolutional layers of the second and first layer groups. However, these lower-layer maps appeared spatially diffuse and consistently underperformed higher-layer counterparts across all tasks except one, as determined by our quantitative explanation analysis using relevance mass accuracy (see ST-2).

**SmoothGrad**
SmoothGrad (Smilkov et al., 2017) was implemented using the Zennit library (v0.5.1) with default settings: a noise level of 0.1 and 20 iterations.

**DeepLift**
DeepLift (Shrikumar et al., 2017) was implemented using the *Captum* library (v0.7.0), a standard framework for model interpretability in PyTorch. We used the mean T1-weighted image computed from 10,000 training subjects as the baseline input, providing a more realistic reference activation than a zero baseline. All other parameters followed Captum's default settings.

**GuidedGradCam**
GuidedGradCAM (Selvaraju et al., 2017) was computed using activations from the final residual block of the last layer group - the same choice used for our main GradCAM variant. We used the implementation provided by the Captum library (v0.7.0).

**Excitation Backprop**

Excitation Backprop (Zhang et al., 2018) was implemented using the Zennit library (v0.5.1) with default parameters.

**Guided Backpropagation**

Guided Backpropagation (Springenberg et al., 2014) was implemented using the Captum library (v0.7.0).

**Input × Gradient**

Input × Gradient (Shrikumar et al., 2017) was implemented using the Captum library (v0.7.0).

**(B4) Evaluation Metrics**

**Relevance Mass Accuracy (RMA)**

To quantitatively assess explanation quality, we used **Relevance Mass Accuracy (RMA)** (Arras et al., 2022). RMA quantifies how well an explanation aligns with ground-truth regions by computing the proportion of relevance mass falling within the ground-truth mask:

$$RMA \; = \; \frac{Relevance\ mass\ inside\ the\ ground-truth\ mask}{Total\ relevance\ mass\ assigned}$$

Intuitively, RMA can be interpreted as the percentage of relevance that is "correctly assigned" to meaningful image regions. Higher RMA values indicate that the explanation better overlaps with areas deemed important for the prediction, as defined by the ground-truth masks.

To improve anatomical alignment of brain atlas–based ground-truth masks, we inverse-warped these masks from nonlinear MNI space to each participant's linear MNI space using deformation fields provided by the UKBB, prior to RMA computation. For cIDP-based tasks (Stage 1 and 2), the masks were additionally dilated by 2 mm to capture boundary voxels that may carry relevant information about a cIDP's shape or volume. In contrast, no dilation was applied for the individual lesion masks (Stage 3), as white matter hyperintensities (WMHs) are primarily characterized by local intensity changes rather than well-defined spatial boundaries.

**True Positive Rate (TPR) / Region of Interest Accuracy**

To complement RMA, we computed region-level accuracy (TPR) for cIDPs (Stage 1). TPR captures the percentage of cases where the ground-truth region was successfully identified by the explanation, providing an interpretable estimate of localization performance. We assigned a "hit" (brain region correctly localized) using the following procedure: All brain regions were ranked by the 99th percentile of explanation intensity within each region's mask, as described in the brain age plausibility analysis (SM-A6). A hit was recorded if the ground-truth region ranked among the top 3 regions. We did not define a hit more strictly (e.g., requiring a top-1 or top-2 rank), as small neighboring regions - such as sulci adjacent to target gyri - sometimes ranked very high and may still convey meaningful information about the target region's boundaries or structure. To separately quantify the extent to which unrelated brain regions were mistakenly highlighted, we introduced the False Positive Rate (FPR; see below).

**False Positive Rate (FPR)**

FPR quantifies how often explanations highlight brain regions unrelated to the ground-truth target. To define a false positive, we first dilated the ground-truth region mask by 2 cm and considered any explanation intensity exceeding the 99th percentile within this dilated mask, but located outside of it, as a false positive. This conservative approach ensured that only clearly spurious relevance was penalized, motivated by our prior finding (SM-A3) suggesting that image regions outside the target mask dilated by 2 cm contain effectively no information about the target label.

**SM-C: Comprehensive Quantitative Benchmark Results**

We provide mean RMA scores for all XAI methods and individual tasks, as well as for grouped tasks (see SM-F2), in ST-2. This table also includes the average overlap scores from the brain age plausibility analysis. Standard deviations across the population for each score are reported in ST-3. To emphasize differences between XAI methods rather than differences across evaluation tasks, all scores displayed in Figures 1.c and 4 were min-max scaled within each row. The scaled values used for these visualizations are available in ST-4 (Figures 1.c) and ST-5 (Figure 4).

True positive rates (TPRs) for all XAI methods and cIDP-based tasks are provided in ST-6. Conversely, false positive rates (FPRs) are reported in ST-7.

To assess the statistical robustness of our core finding - that SmoothGrad outperforms LRP and Grad-CAM - we used our per-task mean RMA and SEMs and created per-XAI-method aggregates scores (SEMs via Gaussian error propagation). We conducted pairwise two-sided t-tests on these aggregated values; SmoothGrad significantly outperformed all LRP variants and Grad-CAM configurations (p < 1e-5). Full results are provided in ST-8.

**SM-D: Qualitative Evaluation and Failure Mode Analysis**

**(D1) Conflicting Explanations Deep Dive (Fig 1a, 2)**

We highlight the critical need for rigorous validation of XAI methods in neuroimaging by demonstrating that different methods can yield contradictory results in a clinically relevant application: identifying neurodegenerative disease markers via brain age explanations in patient subgroups (Schulz et al., 2023).

Brain age prediction involves training machine learning (ML) models to estimate chronological age from neuroimaging data. These models are often interpreted as capturing disease-related changes - such as atrophy or ventricular enlargement - as signs of accelerated aging (Cole & Franke, 2017). This interpretation rests on two key assumptions: (1) that neurodegenerative diseases and normal aging share overlapping neurobiological processes, and (2) that models normally trained on healthy individuals cannot easily disentangle aging effects from disease-related changes (Cole & Franke, 2017; Dinsdale et al., 2021; Feng et al., 2020). Supporting this view, numerous studies report that patients with neurodegenerative conditions are consistently predicted to be older than they are, suggesting that disease-specific alterations are misinterpreted as signs of advanced age (Kaufmann et al., 2019; Lee et al., 2022; Siegel et al., 2025).

XAI can be used to identify disease-related features that models misinterpret as signs of accelerated aging by comparing explanations from patients and matched healthy controls. In theory, explanations for patients should highlight pathological patterns contributing to their older predicted age - patterns that should be absent or less pronounced in healthy controls. We applied this approach to multiple sclerosis (MS), examining differences in brain age explanations between MS patients and propensity score–matched healthy controls. Matching was based on demographic, socioeconomic, and genetic variables: sex, age, education level, household income, the Townsend deprivation index, and genetic principal components, as described in (Schulz et al., 2024).

We trained two 3D ResNet-50 models to predict chronological age from T2-weighted brain MRIs using our age-split protocol (see SM-B2). We excluded subjects who lacked a T2-weighted image. The model architecture and training procedure matched those used in the XAI evaluation (see SM-B1, SM-B2). Using all benchmarked XAI methods (see SM-B3), we generated explanations for both MS patients (N = 142) and their matched healthy controls. To highlight features driving abnormally high age predictions in patients, we retained only the positive components of each explanation - those that increased the predicted age (repeating the experiment with absolute explanations yielded comparable result patterns). We smoothed each explanation map with an 8 mm full-width at half-maximum (FWHM) Gaussian kernel to reduce high-frequency noise. To normalize intensities across subjects, we scaled each map to its 99th percentile.

We trained two 3D ResNet-50 models to predict chronological age from T2-weighted brain MRIs using our age-split protocol (see SM-B2), excluding subjects without T2 scans. Model architecture and training setup matched the XAI evaluation (see SM-B1, SM-B2). We generated explanations for all benchmarked XAI methods (see SM-B3) for MS patients (N = 142) and their matched healthy controls. To highlight features driving abnormally high age predictions in patients, we retained only the positive components of each explanation - those that increased the predicted age. We smoothed each explanation map with an 8 mm full-width at half-maximum (FWHM) Gaussian kernel to reduce high-frequency noise. To normalize intensities across subjects, we scaled each map to its 99th percentile.

To identify disease-specific patterns in the explanations, we performed a mass-univariate analysis using Nilearn's permuted ordinary least squares (OLS) regression (200 permutations). At each voxel, explanation intensity was modeled as the dependent variable, with disease status (MS vs. control) as the main predictor, while controlling for age and sex. Alternative models controlling for predicted age and brain age gap (BAG) as well yielded qualitatively similar results. We computed statistical significance maps ($\alpha$ = 0.05, FWE-corrected) and used them to mask voxelwise effect size maps comparing explanation intensity between patients and controls. This approach highlights brain regions where explanation differences were specifically associated with MS diagnosis.

Strikingly, different XAI methods - applied to the same data and model - produced conflicting conclusions about how the brain age model processed MS-specific disease markers (Figure 1.c, 2, Supplementary Material Figure SM-D.F1, SM-D.F2, SM-D.F3). One group of methods (e.g., LRP, Grad-CAM, and Guided Backpropagation) indicated that MS-associated features - most notably the enlarged lateral ventricles - contributed strongly to abnormally high age predictions in MS patients. This suggests that the model interpreted these anatomical changes as signs of accelerated aging. In contrast, another group of methods (e.g., SmoothGrad, DeepLift, and Input × Gradient) found these same regions to be less influential or even down-weighted in MS patients, implying that the model had learned to discount disease-related alterations when estimating age - potentially disentangling pathological effects from normative aging processes. The model might have learned this apparent disentangling from undiagnosed or prodromal cases present in the training set.

These contradicting interpretations - emerging from the application of different XAI methods to the same model and data - underscore the urgent need for rigorous and standardized validation of XAI techniques in neuroimaging. The substantial disagreement among commonly used methods demonstrates that visual inspection alone is insufficient for assessing explanation quality. Our results highlight the importance of developing objective, domain-relevant evaluation criteria to ensure that XAI outputs are meaningful, reliable, and clinically interpretable.

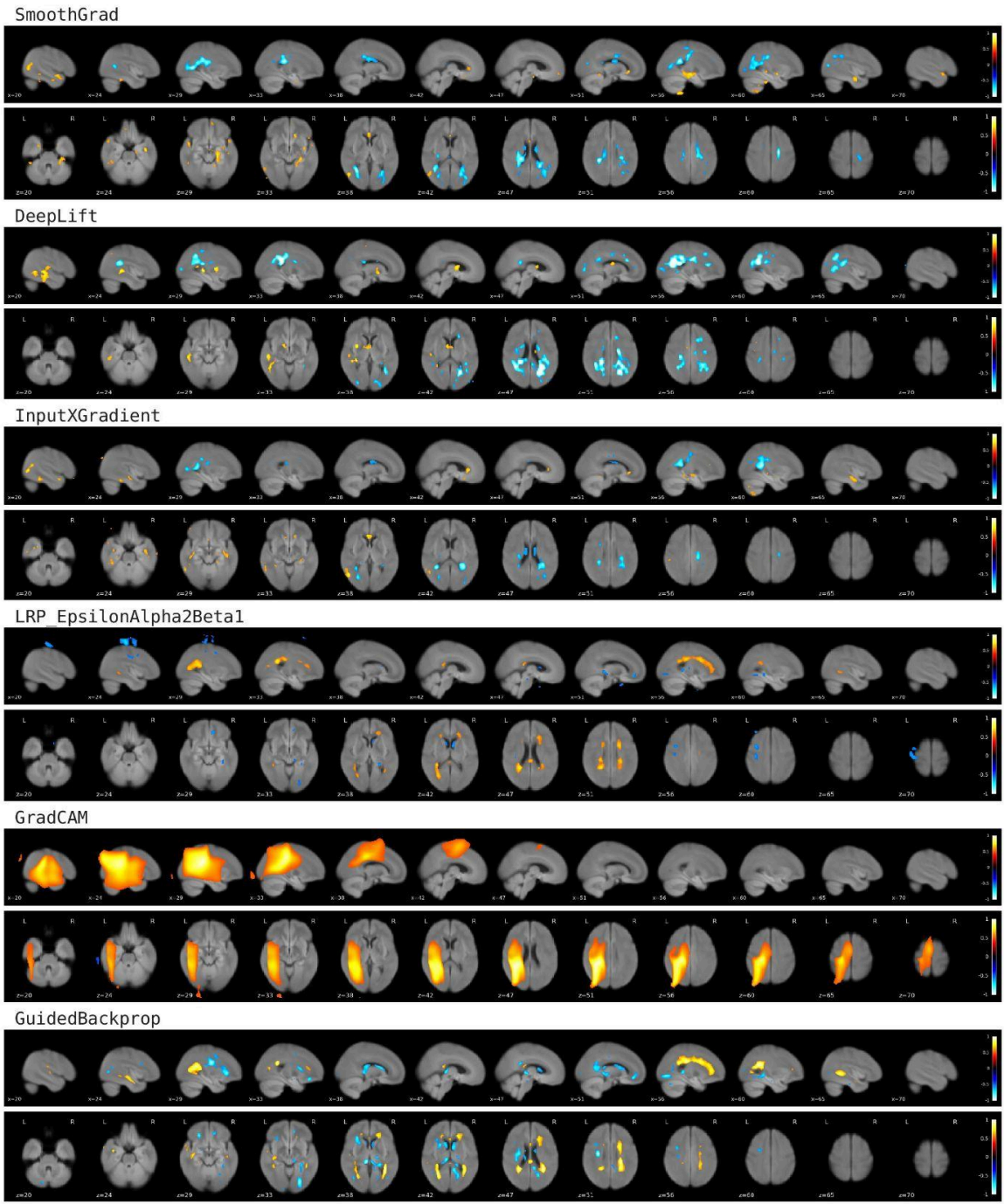

Figure SM-D.F1: Different XAI methods produce conflicting interpretations of how multiple sclerosis (MS) influences brain age model predictions. Each row displays effect size maps comparing explanations for MS patients versus matched healthy controls, masked for statistical significance (α = 0.05, FWE-corrected). Warm colors indicate brain regions that contributed more strongly to age predictions in MS patients, while cool colors indicate reduced contribution relative to controls. Brighter intensities reflect larger effect sizes. The top three methods (SmoothGrad, DeepLift, and Input × Gradient) suggest that regions surrounding the lateral ventricles were down-weighted in MS patients. In contrast, the bottom three methods (LRP, Grad-CAM, and Guided Backpropagation) highlight the same ventricular regions, along with the corpus callosum and cingulate white matter, as highly relevant for predicting age in MS patients.

`SmoothGrad`

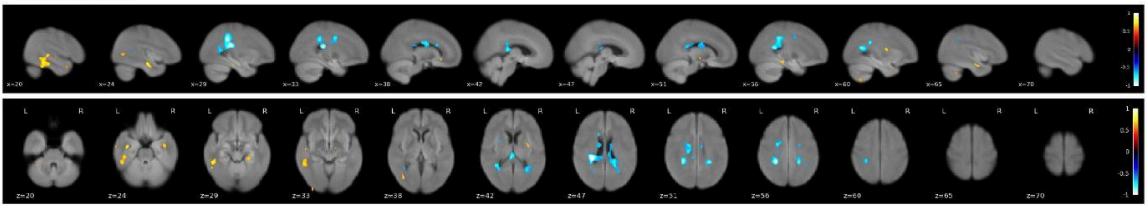

`DeepLift`

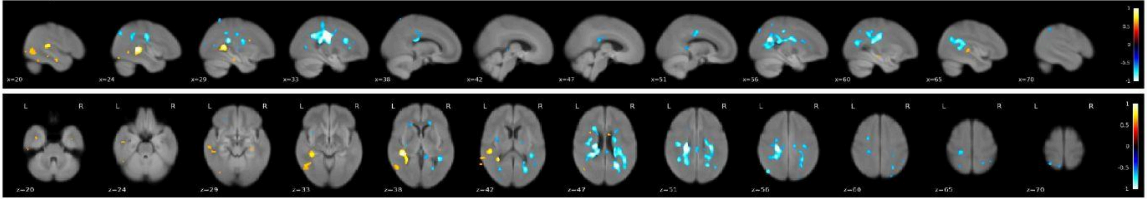

`InputXGradient`

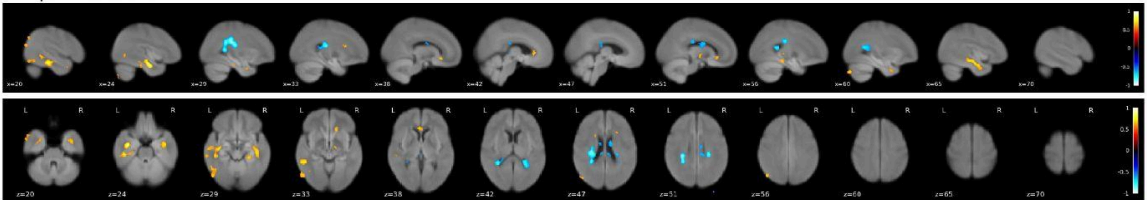

`LRP EpsilonAlpha2Beta1`

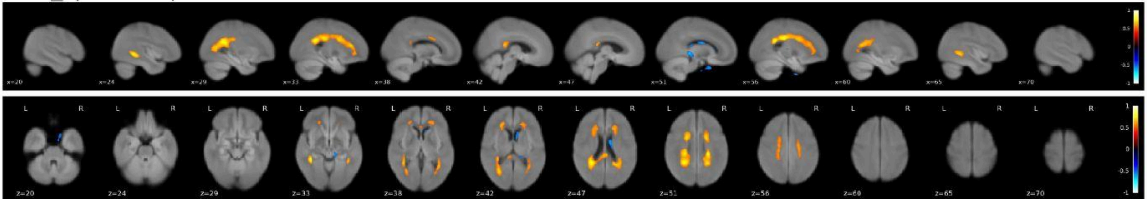

`GradCAM`

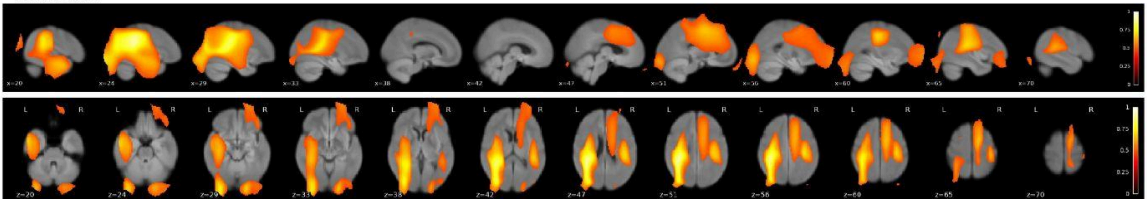

`GuidedBackprop`

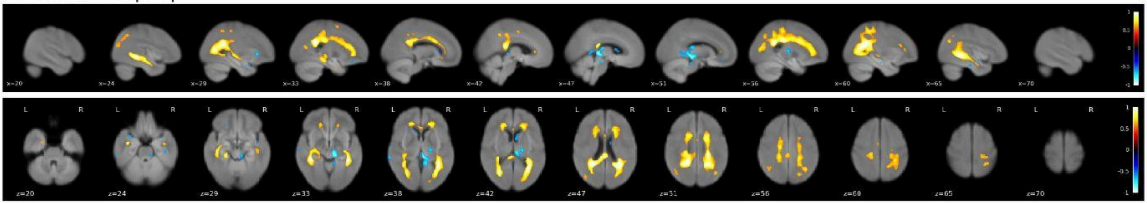

Figure SM-D.F2: Reproduction of Figure SM-D.F1 using a different random model initialization and training batch order.

LRP_EpsilonAlpha2Beta1
MS Patient 0

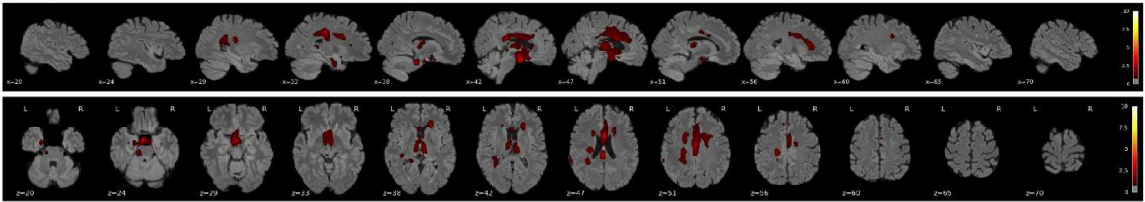

MS Patient 1

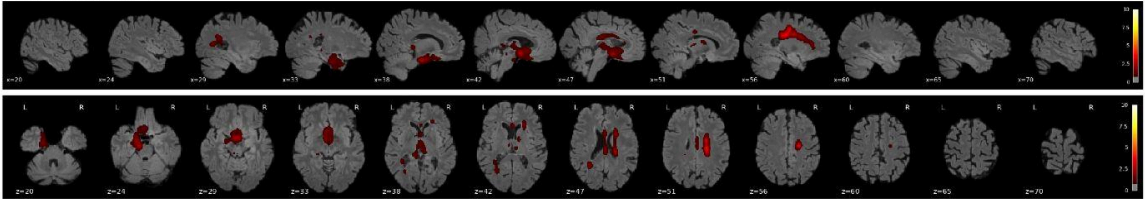

MS Patient 2

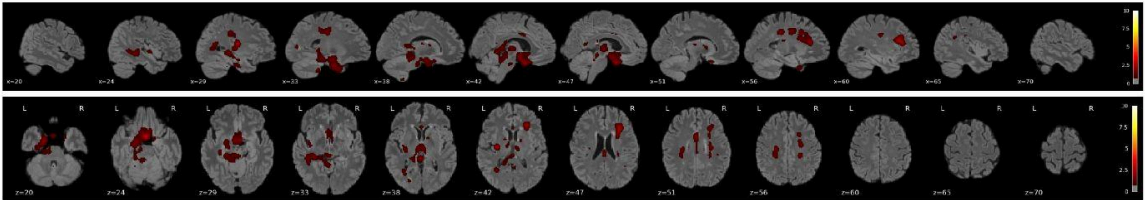

SmoothGrad
MS Patient 0

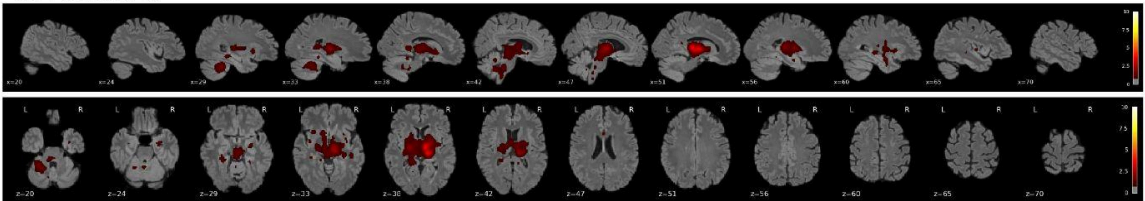

MS Patient 1

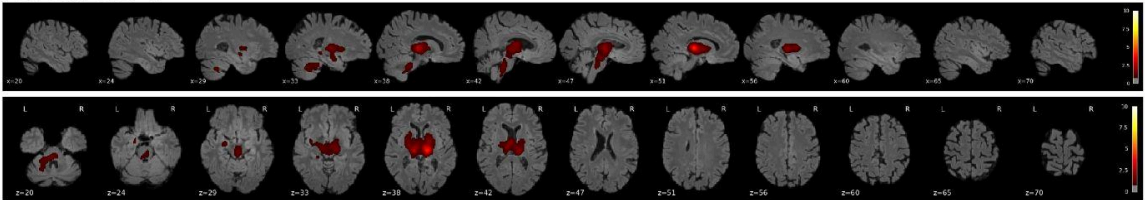

MS Patient 2

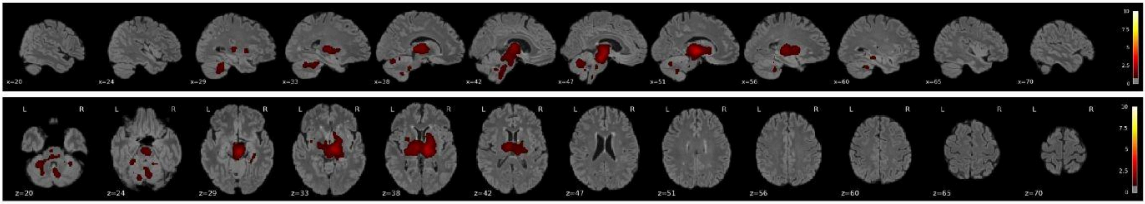

Figure SM-D.F3: Single-subject brain age explanations for MS patients generated using SmoothGrad and LRP. Brighter colors indicate higher feature relevance for the age prediction. Each explanation is visualized with a color scale clipped between 0.75× and 10× the 99th percentile of the respective subject's explanation values (from dark red to white). Notably, SmoothGrad reveals little to no relevance around the lateral ventricles, whereas LRP consistently highlights these regions - underscoring the substantial disagreement between methods in identifying disease-related contributions to the model's age estimates.

## (D2) Extended LRP Artifact Showcase (Fig 3a)

Visual inspection revealed artifacts across all tested LRP variants, including LRP_EpsilonAlpha2Beta1, LRP_EpsilonAlpha2Beta1Flat, LRP_EpsilonPlus, and LRP_EpsilonPlusFlat. See Supplementary Figures SF-1 to SF-4 for both mean and single-subject explanations (brighter colors indicate higher feature relevance; color scales are standardized across explanations) from each LRP variant across cIDP-based tasks. These qualitative observations were supported by high false positive rates (FPR) across a diverse set of cIDP prediction tasks, as shown in SM-D.T1. In contrast, SmoothGrad yielded substantially lower FPRs across all tasks, highlighting the susceptibility of LRP-based methods to spurious attributions outside the verified ground truth regions.

| | LRP_EpsilonAlpha2Beta1 | LRP_EpsilonAlpha2Beta1Flat | LRP_EpsilonPlus | LRP_EpsilonPlusFlat | SmoothGrad |
|---|---|---|---|---|---|
| Mean intensity of Pallidum | 0.28 | 0.11 | 0.45 | 0.25 | 0.01 |
| Mean intensity of Putamen | 0.48 | 0.26 | 0.40 | 0.30 | 0.01 |
| Volume of Caudate | 0.53 | 0.29 | 0.59 | 0.38 | 0.00 |
| Volume of Hippocampus | 0.68 | 0.41 | 0.72 | 0.44 | 0.00 |
| Mean thickness of G-insular-short | 0.98 | 0.86 | 0.99 | 0.94 | 0.04 |
| Mean thickness of G-postcentral | 0.83 | 0.63 | 0.86 | 0.73 | 0.22 |
| Area of G-orbital | 0.32 | 0.26 | 0.35 | 0.23 | 0.05 |
| Area of G-rectus | 0.22 | 0.08 | 0.15 | 0.09 | 0.02 |

Table SM-D.T1: False positive rates (FPR) for four LRP variants and SmoothGrad across eight cIDP tasks. All LRP methods show elevated FPRs, indicating that LRP produces false positive artifacts regardless of the specific composite rule set used.

## (D3) Extended GradCAM Localization Failure Showcase (Fig 3b)

Despite varying the resolution of the activation maps used in Grad-CAM - from the final convolutional layer in the network (layer group 4) to the last convolutional layers in earlier groups (layer groups 3, 2, and 1) - all variants failed to reliably localize ground truth regions across cIDP tasks. See SF-5 to SF-8 for qualitative examples (mean and single-subject explanations; brighter colors indicate higher feature relevance) illustrating Grad-CAM's localization failure across activation layers. Grad-CAM also produced low true positive rates (TPR) regardless of the selected activation layer (Table SM-D.T2), indicating that increasing spatial resolution does not resolve its poor localization performance. For comparison, SmoothGrad achieved consistently high TPR across the same tasks.

|  | GradCAM_l1 | GradCAM_l2 | GradCAM_l3 | GradCAM_l4 | SmoothGrad |
|---|---|---|---|---|---|
| Mean intensity of Pallidum | 0.01 | 0.20 | 0.90 | 0.71 | 0.99 |
| Mean intensity of Putamen | 0.02 | 0.28 | 0.00 | 0.00 | 0.99 |
| Volume of Caudate | 0.05 | 0.27 | 0.08 | 0.84 | 1.00 |
| Volume of Hippocampus | 0.04 | 0.06 | 0.65 | 0.06 | 0.99 |
| Mean thickness of G-insular-short | 0.02 | 0.04 | 0.00 | 0.00 | 0.99 |
| Mean thickness of G-postcentral | 0.02 | 0.02 | 0.00 | 0.14 | 1.00 |
| Area of G-orbital | 0.03 | 0.09 | 0.04 | 0.68 | 0.99 |
| Area of G-rectus | 0.01 | 0.13 | 0.18 | 0.61 | 0.99 |

Table SM-D.T2: True positive rates (TPR) for Grad-CAM applied at different layers - the last convolutional layer in the network (GradCAM_l4), and the last convolutional layers of layer groups 3, 2, and 1 (GradCAM_l3, GradCAM_l2, GradCAM_l1) - as well as for SmoothGrad, across eight cIDP tasks. Grad-CAM consistently underperformed across all layers, suggesting that its failure to localize ground truth targets is not attributable to the resolution of the activation maps. SmoothGrad is included as a reference method, demonstrating robust and reliable localization performance.

## (D4) Extended SmoothGrad Localization Success Showcase (Fig 5)

SmoothGrad achieved successful localization of ground truth regions across cIDP tasks. SF-9 provides extended qualitative examples (mean and single-subject explanations; brighter colors indicate higher feature relevance), further illustrating SmoothGrad's strong alignment with relevant features across all evaluated tasks.

## (D5) Extended Discussion of Domain Mismatch Effects on LRP and GradCAM

The results of our domain mismatch investigation suggest that the failure of LRP and GradCAM stems from these methods being designed and optimized for natural images. In its canonical form, GradCAM relies on activations from the last convolutional layer, implicitly assuming a certain structure in the input data and model representations (Selvaraju et al., 2017). Specifically, it assumes that entities of interest - typically canonical objects - are composed through a hierarchy of features, where simple elements (e.g., edges, shapes) combine into more complex patterns, culminating in object-level concepts represented in higher layers. Based on this assumption, GradCAM uses higher-layer activations to localize where specific objects of interest appear in the input image. However, in neuroimaging, features of interest often lack this compositional hierarchy, calling into question whether CNN representations in this domain follow a similar structure. For example, white matter lesions - a common disease marker - can be simply described as voxels with abnormally high intensities, rather than complex, composite patterns. Consequently, relying on high-layer activations to locate these low-level features may be ineffective. Conversely, although lower-layer activations might better capture simpler features, they tend to be noisy and fragmented, making it difficult for GradCAM to generate meaningful or interpretable localization maps.

Moreover, GradCAM implicitly assumes that objects of interest have a certain spatial scale, since activation maps from higher convolutional layers are low-resolution. This means that objects must be sufficiently large to produce meaningful and interpretable heatmaps. In contrast, features relevant to neuroimaging are often subtle and spatially small, such as tiny lesions or subtle structural variations. These fine-grained features may be lost or blurred in the coarse resolution of higher-layer activations, further limiting GradCAM's suitability for interpreting neuroimaging models.

Unlike GradCAM, LRP generates high-resolution heatmaps by propagating relevance scores back to the input space, which in principle makes it better suited for explaining fine-grained features in neuroimaging. Initially, recommendations for selecting propagation rules and parameters were largely based on qualitative observations (Bach et al., 2015), with best-practice guidelines emerging over time and later validated quantitatively on natural image benchmarks (Kohlbrenner et al., 2020). Montavon et al. (2019) illustrate this evolution: explanations generated using the most basic LRP rule (LRP_0, equivalent to I×G) tend to be noisy and poorly delineate objects, whereas explanations produced by composite best-practice rule sets are sparser and delineate object edges clearly. Although these sparse, edge-focused explanations are intuitive and visually appealing, tuning LRP in this way may have introduced a bias toward natural-image features. Such a bias could underlie LRP's shortcomings in neuroimaging, where features of interest often lack sharp, well-defined boundaries. For example, subcortical volumes are defined by spatial intensity gradients rather than distinct edges. As a result, models in this domain may rely on features that differ substantially from those in natural images. LRP, with rule sets tuned to emphasize edges, may therefore highlight irrelevant high-contrast transitions instead of the smooth, diffuse features that actually drive the model's prediction.

**SM-E: Domain Mismatch Investigation (Fig 4)**

**(E1) Natural Image Benchmark Setup**

To investigate whether the poor performance of the most commonly used XAI techniques in neuroimaging - specifically LRP and Grad-CAM - stems from their design being tailored to natural image tasks, we evaluated all benchmarked XAI methods on the ImageNet dataset, using object segmentation masks as a proxy for ground-truth explanations. These masks provide pixel-level annotations of the main object corresponding to an image's class label, offering a reasonable approximation for image regions relevant to a model's classification decision.

Specifically, we used segmentation masks (N = 12,419) for the ImageNet validation and test sets provided by Gao et al. (2023) as ground-truth explanations. To closely mirror the architecture used in our neuroimaging experiments, we use a 2D ResNet-50 initialized with the default ImageNet pre-trained weights (IMAGENET1K_V2) provided by the PyTorch model zoo[3]. Images were preprocessed following standard ImageNet conventions: resized to 256×256 pixels, center-cropped to 224×224, and normalized using mean = [0.485, 0.456, 0.406] and standard deviation (STD) = [0.229, 0.224, 0.225]. Explanations were computed for the same set of XAI methods and settings as in the neuroimaging benchmark (see SM-B3), except for DeepLift, where a zero baseline was used instead of a mean image. All explanations targeted the logit corresponding to the true class label.

Explanation postprocessing was as follows: to focus on spatial localization rather than channel-specific attribution, we summed relevance across color channels - except for Grad-CAM, which is inherently single-channel. Negative relevance values were then discarded, as our evaluation focused on features that contributed positively to the true class logit, in line with using class-specific object masks as ground truth. Finally, we applied a percentile-based threshold to remove low-magnitude noise and improve correspondence with annotated objects; after testing multiple thresholds (0th, 70th, 80th, 90th, and 95th percentiles), we used the 95th percentile for all evaluations, which yielded RMA scores nearly identical to using the optimal threshold for each XAI method and object category (see SM-F3).

---

[3] https://docs.pytorch.org/vision/main/models/generated/torchvision.models.resnet50.html

Segmentation mask processing was performed to align the explanation ground truth with the preprocessed inputs and evaluation goals. Masks were first resized to 256 × 256 and center-cropped to 224 × 224 to match the input image dimensions. For each image, only the mask corresponding to the labeled object class was retained, with unrelated object masks discarded. To allow boundary pixels to be included in the ground truth, each mask was dilated by 3 pixels. Finally, to ensure that localization remained a meaningful challenge for the XAI methods, we excluded masks that covered more than half the image area - yielding a final set of 7,769 masks for evaluation.

Finally, we computed RMA scores using the post-processed explanations and the corresponding segmentation masks. To assess performance across diverse object categories, we report average RMA scores and STDs for images grouped into semantic supercategories (see SM-E2). The mapping between ImageNet classes and semantic supercategories is provided in ST-9, and the number of usable segmentation masks per category is listed in ST-10.

### (E2) Cross-Domain Performance Comparison

Our cross-domain comparison suggests that the suitability of different XAI methods depends strongly on the data domain. The full results of this analysis are reported in ST-11 (RMA scores), with population-level STDs provided in ST-12. To facilitate method-wise comparisons in Figure 4, RMA scores were row-wise min-max scaled, emphasizing relative differences between XAI methods rather than between semantic categories. These scaled values are available in ST-5.

### SM-F: Method Sensitivity, Robustness, and Post-Processing

**(F1) LRP Sensitivity Analysis:** [dropped from here, moved to separate manuscript]

### (F2) Robustness of XAI methods under varying task characteristics

To systematically assess the robustness of XAI methods under varying task characteristics, we analyzed how explanation fidelity varies with key task properties: model prediction accuracy, target size, the presence of distributed predictive information, target type (e.g., subcortical intensities, subcortical volumes, cortical thicknesses, cortical areas), and the presence of high-contrast patterns. Using RMA as our evaluation metric, we grouped tasks according to these properties and computed average RMA scores within each group. This analysis allowed us to assess whether the performance of different explanation methods remains consistent across a range of neuroimaging scenarios, highlighting how task characteristics shape explanation reliability.

Specifically, to assess the impact of spatial distribution, we grouped together the artificial disease tasks (Stage 2) and lesion load prediction (Stage 3), as these rely on XAI methods capturing relevance distributed across multiple brain regions. To evaluate the role of target size, we categorized cIDPs by modality-specific size into "small" and "large" targets (target region sizes in ST-13), ensuring that each group included only one target per modality to avoid confounding size with modality. Brainstem and ventricular targets were excluded from both size groups due to their extreme anatomical characteristics (e.g., disproportionately large size and edge-specific nature).

Similarly, to examine the role of prediction accuracy, we divided cIDP and artificial disease tasks into "high-accuracy" and "low-accuracy" groups based on test-set mean absolute error (for cIDPs) and classification accuracy (for artificial diseases; see ST-1). We also defined a "high-contrast" target group - characterized by well-delineated anatomical structures with stark intensity boundaries - which included the brainstem and lateral ventricles. Finally, we introduced target type as a categorical grouping reflecting the nature of the predicted phenotype, covering subcortical intensities, subcortical volumes, cortical thicknesses, and cortical areas.

For each of these groupings, we computed average RMA scores across constituent tasks to evaluate whether the reliability of explanation methods varies systematically with task characteristics. Rather than focusing on any specific evaluation task, our analysis aimed to characterize broader patterns of explanation performance across diverse neuroimaging contexts.

Final task groupings:

- **Subcortical intensities**: Mean intensity of Pallidum, Mean intensity of Putamen
- **Subcortical volumes**: Volume of Hippocampus, Volume of Caudate
- **Cortical thicknesses**: Mean thickness of G-insular-short, Mean thickness of G-postcentral
- **Cortical areas**: Area of G-orbital, Area of G-rectus
- **High contrast targets**: Volume of Brain Stem, Volume of Lateral Ventricle
- **Small targets**: Mean intensity of Pallidum, Volume of Caudate, Mean thickness of G-insular-short, Area of G-rectus
- **Large targets**: Mean intensity of Putamen, Volume of Hippocampus, Mean thickness of G-postcentral, Area of G-orbital
- **Low-accuracy targets**: Volume of Hippocampus, Mean thickness of G-insular-short, Volume of Lateral Ventricle, high Mean thickness of G-postcentral & low Volume of Hippocampus
- **High-accuracy targets**: Volume of Caudate, Mean thickness of G-postcentral, Area of G-orbital, Volume of Brain Stem, high Area of G-rectus & low Volume of Caudate
- **Distributed targets**: Artificial disease 1, Artificial disease 2, Lesion load

## (F3) Post-Processing Effects

To prepare explanation maps for evaluation, we applied several postprocessing steps designed to enhance interpretability and ensure comparability across methods and subjects. First, we took the absolute value of the attributions. Given the nature of our ground-truth targets, we are not interested in whether a feature increases or decreases the model's output - rather, we aim to capture what drives the prediction overall, as our reference masks reflect the presence of relevant signal, not its directionality. To reduce noise and emphasize spatially coherent patterns, we applied spatial smoothing using a 4 mm full-width at half-maximum (FWHM) Gaussian kernel. Next, we scaled each explanation map to its 99th percentile value, which standardizes attribution magnitude and makes explanations more comparable across methods and subjects.

To remove residual noise and improve both localization performance and interpretability, we applied a percentile-based threshold to the explanation maps. For brain images, we evaluated thresholds at the 0th, 80th, 90th, 95th, and 99th percentiles - spanning a range from no cutoff (0th), to a minimally visible cutoff (80th), to higher thresholds (e.g., 95th and 99th) that retain only the most prominent attributions. Thresholds above the 99th percentile were found to yield overly sparse explanations that are difficult to interpret visually. For natural images, we followed the same rationale but used slightly lower thresholds - 0th, 70th, 80th, 90th, and 95th percentiles - to reflect the lower proportion of background pixels compared to brain scans.

Based on performance (RMA, brain age marker overlap) across tasks and methods, we selected the 99th percentile threshold for brain images and the 95th percentile for natural images as our default settings, as these yielded the highest scores in nearly all cases. ST-14 (ST-15 for the natural image experiment) reports the scores obtained using the best-performing threshold for each combination of task and XAI method. Notably, the score differences between these optimal per-task thresholds and the fixed defaults (99th for neuroimaging, 95th for ImageNet) are minimal, and the overall performance patterns and method rankings remain consistent - further supporting the robustness of our findings.

**(F4) Architecture Sensitivity Analysis:**

To verify that our results generalize beyond ResNet-50, we repeated our main analyses using a 3D DenseNet-121 architecture (Huang et al., 2017). We kept the training protocol and hyperparameters identical to those used for ResNet. For LRP variants we used DenseNet specific canonization (Pahde et al., 2022). As shown in Supplementary Figure SM-F.F1, DenseNet exhibited the same qualitative pattern of explanation performance across XAI methods and achieved comparable predictive accuracy (ST-16).

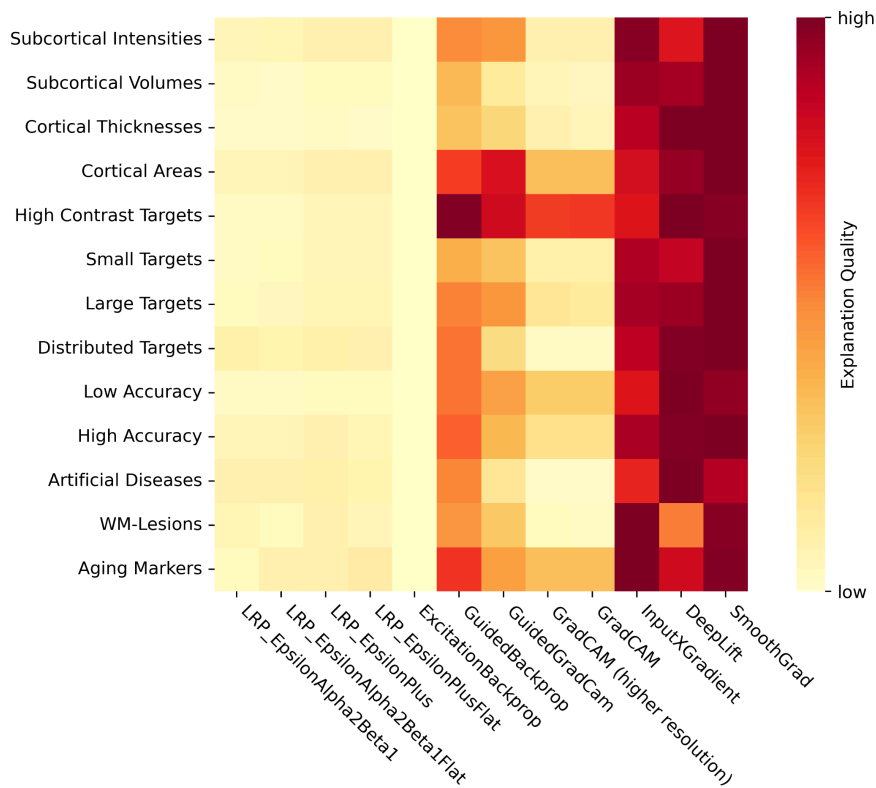

Figure SM-F.F1: Reproducing our main analysis using a 3D DenseNet-121 shows a consistent pattern of XAI performance across model architectures. SmoothGrad performs well across all evaluation stages, while LRP and Grad-CAM fail to achieve high alignment with the ground truth targets. Explanation quality (RMA scores for all rows except *Aging Markers*) is visualized using a color scale from light yellow (low explanation quality) to dark red (high explanation quality).

We provide the mean RMA scores obtained in the reproduction analysis for all XAI methods and individual tasks, as well as for grouped tasks (see SM-F2), in ST-17. This table also reports the average overlap scores from the brain age plausibility analysis. Standard deviations across the population for these reproduction scores are included in ST-18. To highlight differences between XAI methods rather than across evaluation tasks, all scores shown in Figure SM-F.F1 were min-max scaled within each row, analogous to the scaling applied in Figures 1.c and 4. The scaled values used for Figure SM-F.F1 are provided in ST-19.

Reproduction results using the best-performing thresholds for each task and method are reported in ST-20, again showing minimal differences compared to the fixed threshold (99th percentile) results.

**SM-G: XAI Method Usage in Neuroimaging Literature**

To contextualize our benchmark results within current practice, we drew on two recent surveys of explainable AI in medical image analysis to assess the usage frequency of different XAI methods in neuroimaging with deep learning. Specifically, we used *Explainable Artificial Intelligence in Deep Learning-Based Medical Image Analysis* (van der Velden et al., 2022) and *Survey of Explainable Artificial Intelligence Techniques for Biomedical Imaging with Deep Neural Networks* (Nazir et al., 2023). From these sources, we extracted the number of occurrences of each XAI method applied in the brain imaging context. Both reviews indicate that GradCAM (including variants such as CAM or GradCAM++) is the most commonly used method, followed by LRP. Tables with occurrence counts for each method are displayed below.

| Method | Count |
|---|---|
| GradCAM (including CAM) | 23 |
| LRP | 5 |
| Trainable Attention | 3 |
| Backpropagation | 3 |
| Guided Backpropagation | 2 |
| Occlusion Sensitivity | 2 |
| Prediction Difference Analysis | 2 |
| Deconvolution | 1 |
| Multiple Instance Learning | 1 |

Table SM-G.T1: Counts of XAI method usage in brain imaging applications with deep learning, extracted from the 2021 review "Explainable Artificial Intelligence in Deep Learning-Based Medical Image Analysis" (van der Velden et al., 2022). GradCAM (including CAM and GradCAM++ variants) appears most frequently, followed by LRP.

| Method | Count |
|---|---|
| Grad-CAM (including CAM and Grad-CAM++) | 9 |
| LRP | 3 |
| SHAP (SHapley Additive exPlanations) | 2 |
| LIME  (Local Interpretable Model-agnostic Explanations) | 2 |

| Method | Count |
|---|---|
| t-SNE visualization | 1 |
| SmoothGrad | 1 |
| Guided Backpropagation | 1 |
| MSFI visualization | 1 |
| Susceptibility-Weighted Images using Relevance Analysis | 1 |
| Occlusion Sensitivity | 1 |

Table SM-G.T2: Counts of XAI method usage in deep learning-based brain imaging studies, based on the 2023 review "Survey of Explainable Artificial Intelligence Techniques for Biomedical Imaging with Deep Neural Networks" (Nazir et al., 2023). Method frequencies mirror those in the 2021 review, with GradCAM followed by LRP again being the most commonly applied.

# References

Abrol, A., Bhattarai, M., Fedorov, A., Du, Y., Plis, S., Calhoun, V., & Alzheimer's Disease Neuroimaging Initiative. (2020). Deep residual learning for neuroimaging: An application to predict progression to Alzheimer's disease. *Journal of Neuroscience Methods*, *339*(108701), 108701.

Anders, C. J., Neumann, D., Samek, W., Müller, K.-R., & Lapuschkin, S. (2021). Software for dataset-wide XAI: From local explanations to global insights with Zennit, CoRelAy, and ViRelAy. In *arXiv [cs.LG]*. arXiv. http://arxiv.org/abs/2106.13200

Arras, L., Osman, A., & Samek, W. (2022). CLEVR-XAI: A benchmark dataset for the ground truth evaluation of neural network explanations. *An International Journal on Information Fusion*, *81*, 14–40.

Bach, S., Binder, A., Montavon, G., Klauschen, F., Müller, K.-R., & Samek, W. (2015). On pixel-wise explanations for non-linear classifier decisions by layer-wise relevance propagation. *PloS One*, *10*(7), e0130140.

Bethlehem, R. A. I., Seidlitz, J., White, S. R., Vogel, J. W., Anderson, K. M., Adamson, C., Adler, S., Alexopoulos, G. S., Anagnostou, E., Areces-Gonzalez, A., Astle, D. E., Auyeung, B., Ayub, M., Bae, J., Ball, G., Baron-Cohen, S., Beare, R., Bedford, S. A., Benegal, V., … Alexander-Bloch, A. F. (2022). Brain charts for the human lifespan. *Nature*, *604*(7906), 525–533.

Chatterjee, S., Nizamani, F. A., Nürnberger, A., & Speck, O. (2022). Classification of brain tumours in MR images using deep spatiospatial models. *Scientific Reports*, *12*(1), 1505.

Cole, J. H., & Franke, K. (2017). Predicting Age Using Neuroimaging: Innovative Brain Ageing Biomarkers. *Trends in Neurosciences*, *40*(12), 681–690.

Debette, S., & Markus, H. S. (2010). The clinical importance of white matter hyperintensities on brain magnetic resonance imaging: systematic review and meta-analysis. *BMJ (Clinical Research Ed.)*, *341*(jul26 1), c3666.

Destrieux, C., Fischl, B., Dale, A., & Halgren, E. (2010). Automatic parcellation of human cortical gyri and sulci using standard anatomical nomenclature. *NeuroImage*, *53*(1), 1–15.

Dinsdale, N. K., Bluemke, E., Smith, S. M., Arya, Z., Vidaurre, D., Jenkinson, M., & Namburete, A. I. L. (2021). Learning patterns of the ageing brain in MRI using deep convolutional networks. *NeuroImage*, *224*, 117401.

Di Stadio, A., Dipietro, L., Ralli, M., Meneghello, F., Minni, A., Greco, A., Stabile, M. R., & Bernitsas, E. (2018). Sudden hearing loss as an early detector of multiple sclerosis: a systematic review. *European Review for Medical and Pharmacological Sciences*, *22*(14), 4611–4624.

Feng, X., Lipton, Z. C., Yang, J., Small, S. A., Provenzano, F. A., Initiative, A. D. N., & Initiative, F. L. D. N. (2020). Estimating brain age based on a uniform healthy population with deep learning and structural magnetic resonance imaging. *Neurobiology of Aging*, *91*, 15–25.

Fischl, B., Salat, D. H., Busa, E., Albert, M., Dieterich, M., Haselgrove, C., van der Kouwe, A., Killiany, R., Kennedy, D., Klaveness, S., Montillo, A., Makris, N., Rosen, B., & Dale, A. M. (2002). Whole brain segmentation: automated labeling of neuroanatomical structures in the human brain. *Neuron*, *33*(3), 341–355.

Gao, S., Li, Z.-Y., Yang, M.-H., Cheng, M.-M., Han, J., & Torr, P. (2023). Large-scale unsupervised semantic segmentation. *IEEE Transactions on Pattern Analysis and Machine Intelligence*, *45*(6), 7457–7476.

Griffanti, L., Zamboni, G., Khan, A., Li, L., Bonifacio, G., Sundaresan, V., Schulz, U. G., Kuker, W., Battaglini, M., Rothwell, P. M., & Jenkinson, M. (2016). BIANCA (Brain Intensity AbNormality Classification Algorithm): A new tool for automated segmentation of white matter hyperintensities. *NeuroImage*, *141*, 191–205.

Hara, K., Kataoka, H., & Satoh, Y. (2018). Can spatiotemporal 3d cnns retrace the history of 2d cnns and imagenet? *Proceedings of the IEEE Conference on Computer Vision and Pattern Recognition*, 6546–6555.

Huang, G., Liu, Z., Van Der Maaten, L., & Weinberger, K. Q. (2017, July). Densely connected convolutional networks. *2017 IEEE Conference on Computer Vision and Pattern Recognition*

*(CVPR)*. 2017 IEEE Conference on Computer Vision and Pattern Recognition (CVPR), Honolulu, HI. https://doi.org/10.1109/cvpr.2017.243

Jonsson, B. A., Bjornsdottir, G., Thorgeirsson, T. E., Ellingsen, L. M., Walters, G. B., Gudbjartsson, D. F., Stefansson, H., Stefansson, K., & Ulfarsson, M. O. (2019). Brain age prediction using deep learning uncovers associated sequence variants. *Nature Communications*, *10*(1), 5409.

Kang, X., Wang, D., Lin, J., Yao, H., Zhao, K., Song, C., Chen, P., Qu, Y., Yang, H., Zhang, Z., Zhou, B., Han, T., Liao, Z., Chen, Y., Lu, J., Yu, C., Wang, P., Zhang, X., Li, M., ... Multi-Center Alzheimer's Disease Imaging (MCADI) Consortium. (2024). Convergent neuroimaging and molecular signatures in mild cognitive impairment and Alzheimer's disease: A data-driven meta-analysis with N = 3,118. *Neuroscience Bulletin*, *40*(9), 1274–1286.

Kaufmann, T., van der Meer, D., Doan, N. T., Schwarz, E., Lund, M. J., Agartz, I., Alnæs, D., Barch, D. M., Baur-Streubel, R., & Bertolino, A. (2019). Common brain disorders are associated with heritable patterns of apparent aging of the brain. *Nature Neuroscience*, *22*(10), 1617–1623.

Kingma, D. P., & Ba, J. (2014). Adam: A method for stochastic optimization. *arXiv Preprint arXiv:1412.6980*.

Kohlbrenner, M., Bauer, A., Nakajima, S., Binder, A., Samek, W., & Lapuschkin, S. (2020, July). Towards best practice in explaining neural network decisions with LRP. *2020 International Joint Conference on Neural Networks (IJCNN)*. 2020 International Joint Conference on Neural Networks (IJCNN), Glasgow, United Kingdom. https://doi.org/10.1109/ijcnn48605.2020.9206975

Kolbeinsson, A., Filippi, S., Panagakis, Y., Matthews, P. M., Elliott, P., Dehghan, A., & Tzoulaki, I. (2020). Accelerated MRI-predicted brain ageing and its associations with cardiometabolic and brain disorders. *Scientific Reports*, *10*(1), 19940.

Lee, J., Burkett, B. J., Min, H.-K., Senjem, M. L., Lundt, E. S., Botha, H., Graff-Radford, J., Barnard, L. R., Gunter, J. L., Schwarz, C. G., Kantarci, K., Knopman, D. S., Boeve, B. F., Lowe, V. J., Petersen, R. C., Jack, C. R., Jr, & Jones, D. T. (2022). Deep learning-based brain age prediction in normal aging and dementia. *Nature Aging*, *2*(5), 412–424.

Leonardsen, E. H., Peng, H., Kaufmann, T., Agartz, I., Andreassen, O. A., Celius, E. G., Espeseth, T., Harbo, H. F., Høgestøl, E. A., Lange, A.-M. de, Marquand, A. F., Vidal-Piñeiro, D., Roe, J. M., Selbæk, G., Sørensen, Ø., Smith, S. M., Westlye, L. T., Wolfers, T., & Wang, Y. (2022). Deep neural networks learn general and clinically relevant representations of the ageing brain. *NeuroImage*, *256*(119210), 119210.

Nazir, S., Dickson, D. M., & Akram, M. U. (2023). Survey of explainable artificial intelligence techniques for biomedical imaging with deep neural networks. *Computers in Biology and Medicine*, *156*(106668), 106668.

Pahde, F., Yolcu, G. Ü., Binder, A., Samek, W., & Lapuschkin, S. (2022). Optimizing explanations by network canonization and hyperparameter search. In *arXiv [cs.CV]*. arXiv. http://arxiv.org/abs/2211.17174

Peng, H., Gong, W., Beckmann, C. F., Vedaldi, A., & Smith, S. M. (2021). Accurate brain age prediction with lightweight deep neural networks. *Medical Image Analysis*, *68*(101871), 101871.

Schulz, M.-A., Hetzer, S., Eitel, F., Asseyer, S., Meyer-Arndt, L., Schmitz-Hübsch, T., Bellmann-Strobl, J., Cole, J. H., Gold, S. M., Paul, F., Ritter, K., & Weygandt, M. (2023). Similar neural pathways link psychological stress and brain-age in health and multiple sclerosis. *iScience*, *26*(9), 107679.

Schulz, M.-A., Siegel, N. T., & Ritter, K. (2024). Beyond accuracy: Refining brain-age models for enhanced disease detection. In *bioRxiv* (p. 2024.03. 28.587212). https://doi.org/10.1101/2024.03.28.587212

Selvaraju, R. R., Cogswell, M., Das, A., Vedantam, R., Parikh, D., & Batra, D. (2017, October). Grad-CAM: Visual explanations from deep networks via gradient-based localization. *2017 IEEE International Conference on Computer Vision (ICCV)*. 2017 IEEE International Conference on Computer Vision (ICCV), Venice. https://doi.org/10.1109/iccv.2017.74

Shrikumar, A., Greenside, P., & Kundaje, A. (2017). Learning important features through propagating activation differences. In *arXiv [cs.CV]*. arXiv. http://arxiv.org/abs/1704.02685

Siegel, N. T., Kainmueller, D., Deniz, F., Ritter, K., & Schulz, M.-A. (2025). Do transformers and CNNs

learn different concepts of brain age? *Human Brain Mapping*, *46*(8). https://doi.org/10.1002/hbm.70243

Smilkov, D., Thorat, N., Kim, B., Viégas, F., & Wattenberg, M. (2017). SmoothGrad: removing noise by adding noise. In *arXiv [cs.LG]*. arXiv. http://arxiv.org/abs/1706.03825

Smith, L. N., & Topin, N. (2019). Super-convergence: very fast training of neural networks using large learning rates. In T. Pham (Ed.), *Artificial Intelligence and Machine Learning for Multi-Domain Operations Applications*. SPIE. https://doi.org/10.1117/12.2520589

Springenberg, J. T., Dosovitskiy, A., Brox, T., & Riedmiller, M. (2014). Striving for simplicity: The all convolutional net. In *arXiv [cs.LG]*. arXiv. http://arxiv.org/abs/1412.6806

van der Velden, B. H. M., Kuijf, H. J., Gilhuijs, K. G. A., & Viergever, M. A. (2022). Explainable artificial intelligence (XAI) in deep learning-based medical image analysis. *Medical Image Analysis*, *79*(102470), 102470.

Walhovd, K. B., Westlye, L. T., Amlien, I., Espeseth, T., Reinvang, I., Raz, N., Agartz, I., Salat, D. H., Greve, D. N., & Fischl, B. (2011). Consistent neuroanatomical age-related volume differences across multiple samples. *Neurobiology of Aging*, *32*(5), 916–932.

Wilming, R., Budding, C., Müller, K.-R., & Haufe, S. (2022). Scrutinizing XAI using linear ground-truth data with suppressor variables. *Machine Learning*, *111*(5), 1903–1923.

Younis, A., Li, Q., Afzal, Z., Jajere Adamu, M., Bello Kawuwa, H., Hussain, F., & Hussain, H. (2024). Abnormal brain tumors classification using ResNet50 and its comprehensive evaluation. *IEEE Access: Practical Innovations, Open Solutions*, *12*, 78843–78853.

Zhang, J., Bargal, S. A., Lin, Z., Brandt, J., Shen, X., & Sclaroff, S. (2018). Top-down neural attention by excitation backprop. *International Journal of Computer Vision*, *126*(10), 1084–1102.